# A Switching Mechanism for Selecting Environment-Adaptive Personalized Gains in Hearing Aids

**Nafisa Zarrin Tasnim and Nasser Kehtarnavaz**

Department of Electrical and Computer Engineering, The University of Texas at Dallas, Richardson, TX 75080, USA

**ABSTRACT** Personalized amplification gains for hearing aids have been shown to be preferred over prescriptive gains. Since the preferred settings can vary depending on the background audio environment, this paper presents a switching mechanism or system that automatically selects personalized gains for a current background audio environment. To prevent foreground speech from interfering with background environment analysis, a one-class voice activity detector is designed to identify speech activity from speech mixed with noise. A consistency check is conducted to enable distinguishing foreground speech from background speech, preserving babble as a background environment of interest because hearing aid users often struggle to understand speech in background babble. Sound segments without foreground speech are then analyzed by an online discovery method combining Bayesian change-point detection with a memory of previously encountered environments. This discovery method detects changes, recognizes returning environments, and discovers new ones without a predefined number of environments. Experimental results show that three-seed mean macro-averaged F1 scores of 85.53%-91.98% across eight speech/background-corpus pairs covering 36 background noise types were achieved. Furthermore, the discovery method achieved adjusted Rand indices of 0.91 and 0.96 on two families of test streams. In the same-recording-return condition of audio streams combining speech with four background audio environments often encountered by hearing aid users, the switching mechanism selected the correct personalized gain profile after 105 environment changes, recalled the stored gain profile at every return, and made no switches to an incorrect gain profile.

## I. INTRODUCTION

According to the World Health Organization report in 2021 [1], hearing loss affects more than 1.5 billion people worldwide and can have a substantial impact on their communication, social interaction and quality of life. Hearing aids compensate for hearing loss by amplifying sound in a number of frequency bands, and the gains are commonly set by prescriptions such as NAL-NL2 [2] and DSLv5 [3]. However, these prescriptions are designed based on population data and do not fully account for an individual user's hearing preferences in different audio environments [4]. As discussed in previous studies, e.g. [4], [5], [6], [7], personalization techniques address this issue by adjusting the prescription gains according to the hearing preferences of an individual user. Several machine learning-based approaches have been introduced for personalization of hearing aid amplification gains, including offline deep reinforcement learning [8], [9], online maximum-likelihood inverse reinforcement learning [10], [11], [12], [13], and multi-band Bayesian learning [14], [15], [16], [17].

Although these studies have enabled personalization of gains, their reported experiments primarily focus on one background noise environment of babble. Since hearing preferences of an individual hearing aid user may vary in different background noise environments, a single set of personalized gains may not adequately reflect the user's preferences across different audio environments. Building on the switching mechanism presented in [18], this work develops a voice activity detection and an online environment discovery method to identify a current background audio environment and recognize when the same or a similar environment is encountered. The gain profiles associated with these environments can be obtained and stored using the personalization procedures in the previous studies.

Reliable characterization of the background noise environment involves two challenges. First, previous hearing aid sound classifiers and conventional acoustic scene classification methods typically consider predefined listening conditions or environmental categories, e.g. [19], [20], [21], [22], whereas audio environments encountered by an individual user may not be known in advance. Second, foreground speech can interfere with the characterization of the background audio environment because audio features derived from mixed audio may be influenced by speech characteristics [23]. Identifying sound segments without foreground speech can therefore reduce this interference. However, background speech, such as babble in a restaurant or speech from a television, forms part of the audio environment itself. The relevant distinction is consequently between foreground speech and all other audio rather than between speech and non-speech.

Figure 1 illustrates an overview of the developed switching mechanism for environment-adaptive personalized gains. Incoming audio is first processed by a voice activity detection (VAD) module to identify sound segments in which foreground speech is present or absent. Audio segments in which foreground speech is absent are subsequently analyzed by an online environment change detection module to identify transitions and determine whether the current environment is new or has been encountered previously. The identified environment is then used to select its associated personalized gain profile.

To address the above challenges, this work investigates the detection of foreground speech and the online discovery of background environments as two components of a switching mechanism for selecting environment-adaptive personalized gains. The components are evaluated individually, and their integration is demonstrated on continuous audio streams containing speech and changing background audio environments. More specifically, the following contributions are made in this paper:

I. Development of a one-class foreground VAD that combines a fixed bank of speech-in-noise representations with online speaker-consistency verification to select audio for background-environment analysis.

II. Development of an online environment-discovery method that detects changes, recognizes returning environments, and learns two recognition thresholds from the incoming audio without a predefined number of environments.

III. Development of a switching mechanism that uses the identified environments to select and recall associated personalized gain profiles, evaluated on continuous streams containing speech and changing backgrounds.

The rest of this paper is organized as follows. Section II reviews related works on VAD and environmental audio analysis for hearing devices. Section III presents the one-class foreground VAD with speaker-consistency verification, and Section IV presents the online discovery method. Section V evaluates the automatic selection of personalized gain profiles by the developed switching mechanism. The paper is then concluded in Section VI.

## II. RELATED WORKS

This section reviews the two areas underlying the developed switching mechanism. Section II-A reviews voice activity detection, with emphasis on one-class learning. Section II-B reviews environmental audio analysis, including audio environment classification, online discovery and acoustic change detection.

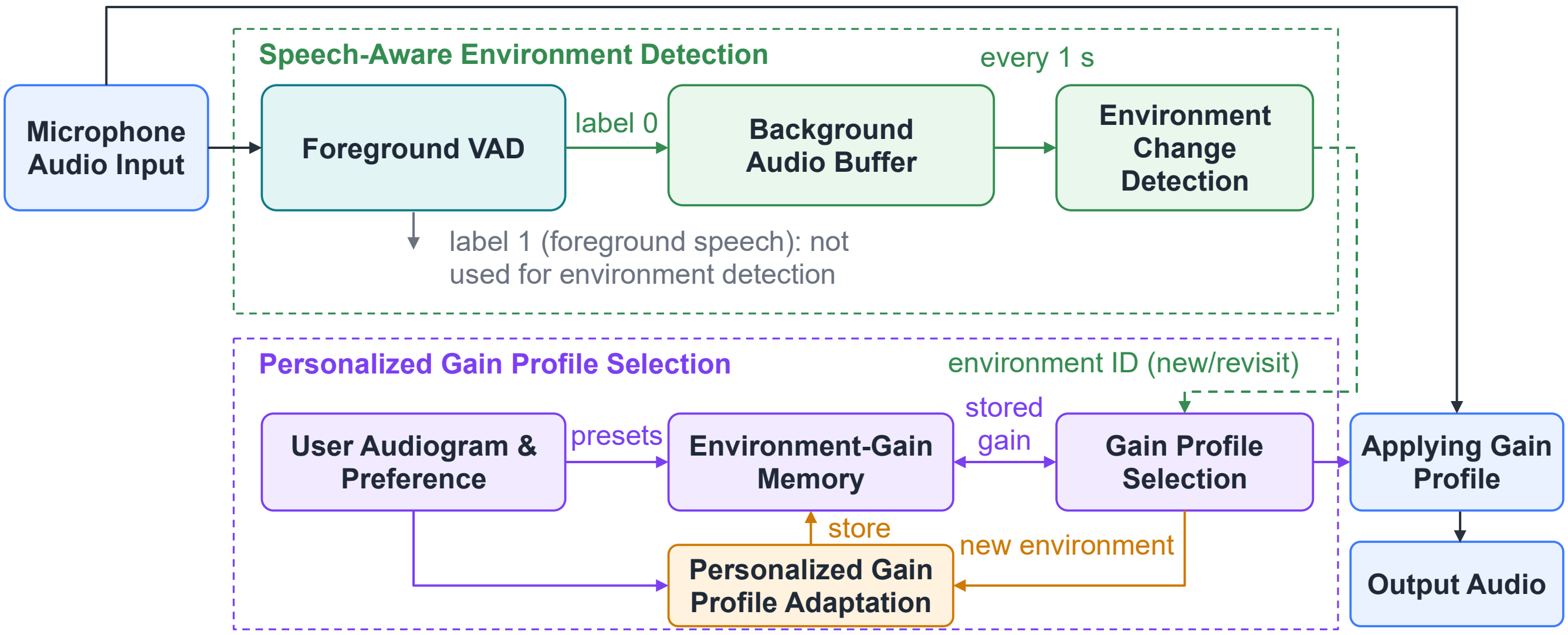

**FIGURE 1.** **Components of the developed switching mechanism for selecting environment-adaptive personalized gains using foreground voice activity detection and online environment discovery.**

## A. VOICE ACTIVITY DETECTION

1) Conventional and Deep Learning-Based VADs

Traditional VAD methods use short-time energy and zero-crossing information for speech endpoint detection [24], statistical likelihood-ratio tests and hangover decisions [25], or long-term spectral information for robustness in adverse conditions [26]. Unsupervised VAD methods such as rVAD [27] combine denoising with segment-based speech decisions to reduce the need for annotated training data. Deep learning methods for VAD use learned acoustic and temporal representations, including deep belief networks [28] and recurrent neural networks [29]. Teacher-student training has also been used to reduce reliance on frame-level annotations [30]. Pretrained general-purpose detectors such as Silero VAD are publicly available [31]. These approaches primarily address speech presence and may detect background talkers along with the foreground speaker, which is undesirable when background speech must remain available for background environment analysis.

2) Self-Supervised Speech and Speaker Representations

Self-supervised speech models provide representations that can be transferred to downstream audio tasks. The wav2vec 2.0 approach uses contrastive prediction over masked latent speech representations [32], HuBERT predicts masked cluster assignments [33], and WavLM incorporates denoising and utterance mixing during pretraining [34]. Speaker-discriminative representations are available through pretrained toolkits such as WeSpeaker [35]. Frozen encoders permit task-specific speech detection and speaker-consistency decisions without retraining these representation models.

3) One-Class Methods

One-class methods characterize the target class when representative observations of alternative classes are limited. Support vector data description [36] and Deep SVDD [37] describe a compact target region, and one-class feature-learning methods seek suitable target representations [38]. A one-class support vector machine estimates the support of a target distribution [39]. Reconstruction-based models have been applied to acoustic novelty detection [40], [41] and anomalous sound detection [42]. For VAD, ORVAE models speech mixed with noisy backgrounds using a one-class residual variational autoencoder [43], while a feature-driven one-class variational autoencoder attempts to detect speech in background noise [44]. In contrast to reconstruction-based scoring, the first stage of the developed detector measures similarity to a fixed bank of speech-in-noise representations.

4) Speaker-Selective and Foreground VAD

Speaker-selective VAD distinguishes target speech from other speech using speaker-related information. Personal VAD uses an enrolled speaker reference to distinguish target speech, non-target speech, and non-speech [45]. Personal VAD 2.0 supports on-device streaming and an enrollment-free mode, although the latter functions as a conventional VAD [46]. Online target-speaker VAD has also estimated speaker embeddings from previous confident detections rather than requiring a fixed reference [47]. More recently, FVAD has investigated enrollment-free foreground speech detection using supervised speaker selectivity and competing-speaker augmentation [48]. The approach in this paper instead combines a fixed one-class speech-in-noise reference bank with an online speaker-consistency memory, without a separate enrollment recording. Table I compares these closely related approaches.

TABLE I
TRAINING AND INFERENCE CHARACTERISTICS OF CLOSELY RELATED VAD APPROACHES

| Approach | Training basis | Speaker reference | Output objective |
|---|---|---|---|
| ORVAE [43] | Speech with noisy backgrounds (one-class) | None | Speech activity |
| Personal VAD [45] | Supervised speaker-dependent labels | Enrolled target embedding | Target speech, other speech, non-speech |
| Personal VAD 2.0 [46] | Supervised; enrolled and unenrolled cases | Enrolled embedding, or none | Target speech, or general speech |
| Online TS-VAD [47] | Supervised speaker-activity labels | Embeddings estimated online | Multi-speaker activity |
| FVAD [48] | Foreground labels with competing speakers | None (implicit) | Foreground speech activity |
| Ours | Speech-in-noise reference bank; frozen encoders | Online speaker memory | Foreground activity |

## B. AUDIO ENVIRONMENT ANALYSIS

1) Environment Classification for Hearing Aids

Smartphone-based applications have supported hearing aid fitting through adjustable compression or gain settings [49] and have provided amplification with noise reduction [50]. Automatic sound classification has also been investigated for selecting hearing aid gains. Previous methods distinguished listening conditions such as clean speech, traffic noise, babble and music using a hidden Markov model or auditory-scene-inspired features [19], [20]. More generally, acoustic scene classification assigns recordings to predefined scenes [21], and convolutional neural networks have improved classification accuracy [22]. An adaptive hearing aid classifier was subsequently developed to split or merge classes as listening conditions changed [51]. Foreground speech poses a further difficulty. Increased foreground speech was shown to reduce scene-classification accuracy in [23], where speech-presence information was combined with noise-floor-based features to improve robustness. Nevertheless, a predefined set of scenes was considered.

2) Online Discovery of Acoustic Environments

Unsupervised approaches seek to identify environmental structure without fixed labels. In [52], a real-time noise classifier used ART2 clustering and added a class when a previously unseen noise environment was encountered; related hearing-enhancement applications incorporated unsupervised noise classification [53], and ART2 clustering was subsequently used for automatic switching of personalized gains [18]. SoundSense identified recurring ambient sounds on mobile devices after separating voice, music and ambient categories [54]. Online speaker-clustering methods similarly assign a new segment to an existing cluster or create a new one from similarity or Bayesian decisions [55], [56]. Density-based methods such as DBSCAN and HDBSCAN identify clusters without fixing their number [57], [58], and DP-means creates clusters through a penalized-distance rule [59]. DenStream maintains clusters over evolving data streams with fading weights [60], although fading can make re-identification after a long absence difficult. Other works segmented audio and then grouped recurring environments, but performed the grouping offline [61], [62]. Pretrained representations such as PANNs and CLAP provide transferable descriptors for environmental analysis [63], [64], [65]. These approaches motivate a persistent environment memory, but observation-level decisions can react to transient events. The approach developed here updates its memory at detected environmental boundaries and learns its matching threshold from the incoming stream itself.

3) Acoustic Change Detection

Acoustic segmentation has been investigated using the Bayesian Information Criterion (BIC) for detecting and clustering speaker, environment, and channel changes [66], and variable-window BIC searches have been used for faster segmentation [67]. Other approaches detect segment boundaries through symmetric Kullback-Leibler distance [68] or combine online speaker change detection with speaker tracking [69], [70]. Among general sequential methods, CUSUM detects accumulated deviations from a reference distribution [71], whereas BOCPD maintains a posterior over the time since the latest change [72]. NEWMA detects changes in high-dimensional streams using exponentially weighted feature averages [73], and acoustic scene changes have also been investigated for low-power hearing devices [74]. A posterior-probability alarm convention for Bayesian change-point models was described in [75]. These methods support the detection of environmental transitions but change detection by itself does not assign a persistent identity when an environment is again encountered or returns. Table II summarizes the environment-analysis approaches most closely related to this paper. Although previous studies have investigated online change detection, the recognition of recurring environments, and foreground speech detection, these aspects have largely been addressed separately. To the best of our knowledge, no published system jointly combines Bayesian online change-point detection, re-identification of recurring background environments using pretrained audio embeddings, and a one-class voice activity detector that selects the audio used for environmental characterization. This is the difference between our approach and what has already been reported in the literature.

TABLE II
CHARACTERISTICS OF CLOSELY RELATED ENVIRONMENT-ANALYSIS APPROACHES

| Approach | Online | Classes | Change detection | Recurring environments / foreground speech |
|---|---|---|---|---|
| Hearing-aid classifiers [19], [20] | Yes | Predefined | No | Predefined classes; speech is a class |
| Song et al. [23] | No | Predefined | No | Predefined scenes; supervised speech-presence features |
| Lamarche et al. [51] | Yes | Adapted (split, merge) | No | Adapted classes; speech is a starting class |
| Alamdari and Kehtarnavaz [52] | Yes | Discovered | No | Recognized by the classifier; VAD selects noise frames |
| SoundSense [54] | Yes | Discovered | No | Recognized; voice handled separately |
| Online diarization clustering [55], [56] | Yes | Discovered | No | Recognized; targets speakers |
| Ellis and Lee [61], [62] | No (post hoc) | Discovered | Yes (BIC) | Recognized by clustering; not addressed |
| Lu and Zhang [70] | Yes | Discovered | Yes | Tracked speaker models; targets speakers |
| Koppelmann and Martin [74] | Yes | Not modeled | Yes | Not recognized; not addressed |
| Ours | Yes | Discovered | Yes | Re-identified with thresholds learned from the stream; foreground VAD selects background audio |

## III. STREAMING ONE-CLASS VOICE ACTIVITY DETECTION WITH SPEAKER-CONSISTENCY VERIFICATION

Within the components of the switching mechanism shown in Fig. 1, this section presents a one-class approach in which the task-specific speech model is fitted using speech-plus-noise examples without fitting a separate background class. The method has two stages: similarity to a fixed speech-in-noise reference bank, followed by a check of speaker consistency against a memory of recent detections, which needs no enrollment recording.

### A. DETECTION OBJECTIVE AND SYSTEM OVERVIEW

The detector identifies activity of a designated foreground speaker so that intervals without this speaker can be used for background-environment analysis. Voices already present in the background, such as restaurant babble, remain part of the environment to be characterized. Each test stream contains utterances from one designated speaker separated by gaps, with background audio added continuously. The resulting stream can be expressed as $x[n] = s[n] + \alpha v[n]$, where $s[n]$ denotes the designated speech component, $v[n]$ denotes the continuous background audio, and $\alpha$ controls the mixture signal-to-noise ratio (SNR). Thus, each stream contains both speech-plus-background and background-only intervals. At each decision time $t$, separated by a frame interval $\Delta_f$, the detector predicts whether the designated speech is present.

Figure 2 provides an illustration of the architecture. Audio is converted to mono at sampling rate $f_s$. Stage 1 extracts speech representations, measures their distance from the positive reference bank, and thresholds the smoothed score. Stage 2 checks speaker consistency using a memory updated from confident observations. A duration filter removes positive runs shorter than $D_{\min}$ decision frames. The feature transform, reference bank, encoders, and thresholds are fixed before testing; only the speaker memory and its centroid change during inference. Parameter values and the basis for their selection are summarized in Table III.

## B. POSITIVE REFERENCE BANK

Speech representations are extracted with the frozen HuBERT Base encoder [33], using the facebook/hubert-base-ls960 checkpoint. Let $h_t$ denote its $d_H$ dimensional hidden state at layer $\ell_H$. Test inference emits blocks of $B_H$ output frames, with $L_H$ frames of past context and $R_H$ frames of future context. Attention remains bidirectional within each supplied chunk; the chunk boundaries limit access to the stream.

A transform fitted to eligible training frames standardizes each feature dimension and applies principal component analysis (PCA) to obtain $d_P$ dimensional vectors. Each transformed vector is normalized to unit Euclidean norm. For decision frame $t$, the elementwise mean $\mu_t$ and standard deviation $\sigma_t$ are computed over frames $t - L_P$ through $t + R_P$, where $L_P$ and $R_P$ specify the pooling context. Their concatenation gives this $2d_P$ dimensional segment representation

$$z_{\mathrm{V},t} = \frac{[\mu_t; \sigma_t]}{\|[\mu_t; \sigma_t]\|_2} \tag{1}$$

The reference bank contains up to $N_{\mathrm{ref}}$ segment vectors extracted from speech-plus-noise mixtures, and the PCA transform is fitted using up to $N_P$ eligible mixture frames. The corresponding clean recordings are used only to identify suitable speech-active centers and to compute local SNR. A center is labeled as speech-active when its clean-reference frame energy exceeds both the utterance's $p_E$th-percentile frame energy minus $\Delta_E$ dB and an absolute floor $E_{\min}$. Positive label runs shorter than $n_{\mathrm{lab}}$ frames are removed. Bank centers must also have local mixture SNR of at least $S_B$, and at least a fraction $\eta_P$ of the frames in their pooling interval must be labeled speech-active. The bank draws equal numbers of centers from the training background and SNR combinations; a combination with fewer eligible centers contributes all of them. The stored vectors pool features from the noisy audio around these centers; noise remains present, and the pooling interval may include background-only frames.

The transform and bank are fitted from offline training features and retained for bounded-context inference. Let $N_t$ contain the $k_{\mathrm{NN}}$ bank vectors with the highest cosine similarity to $z_{\mathrm{V},t}$. Because the query and bank vectors have unit norm, the distance score is

$$d_t = 1 - \frac{1}{k_{\mathrm{NN}}} \sum_{b \in N_t} z_{\mathrm{V},t}{}^{\mathrm{T}} b \tag{2}$$

With the single nearest neighbor of the final configuration ($k_{\mathrm{NN}} = 1$; Table III), this score is the cosine distance to the closest bank vector. The score is smoothed over $M_{\mathrm{s}}$ preceding and $M_{\mathrm{s}}$ following frames as follows:

$$\bar{d}_t = \frac{1}{2M_{\mathrm{s}} + 1} \sum_{j=-M_{\mathrm{s}}}^{M_{\mathrm{s}}} d_{t+j} \tag{3}$$

with edge replication at recording boundaries. Stage 1 accepts a frame when $\bar{d}_t \le \tau_a$. The acceptance threshold $\tau_a$ and the stricter memory-admission threshold $\tau_u$ are the $q_a$ and $q_u$ quantiles of eligible positive calibration scores, respectively, with $q_u < q_a$, computed with equal total weight for each calibration SNR. Both are fitted on held-out positive calibration regions. Eligible calibration centers satisfy the selection label defined in Section III-B; note that the local-

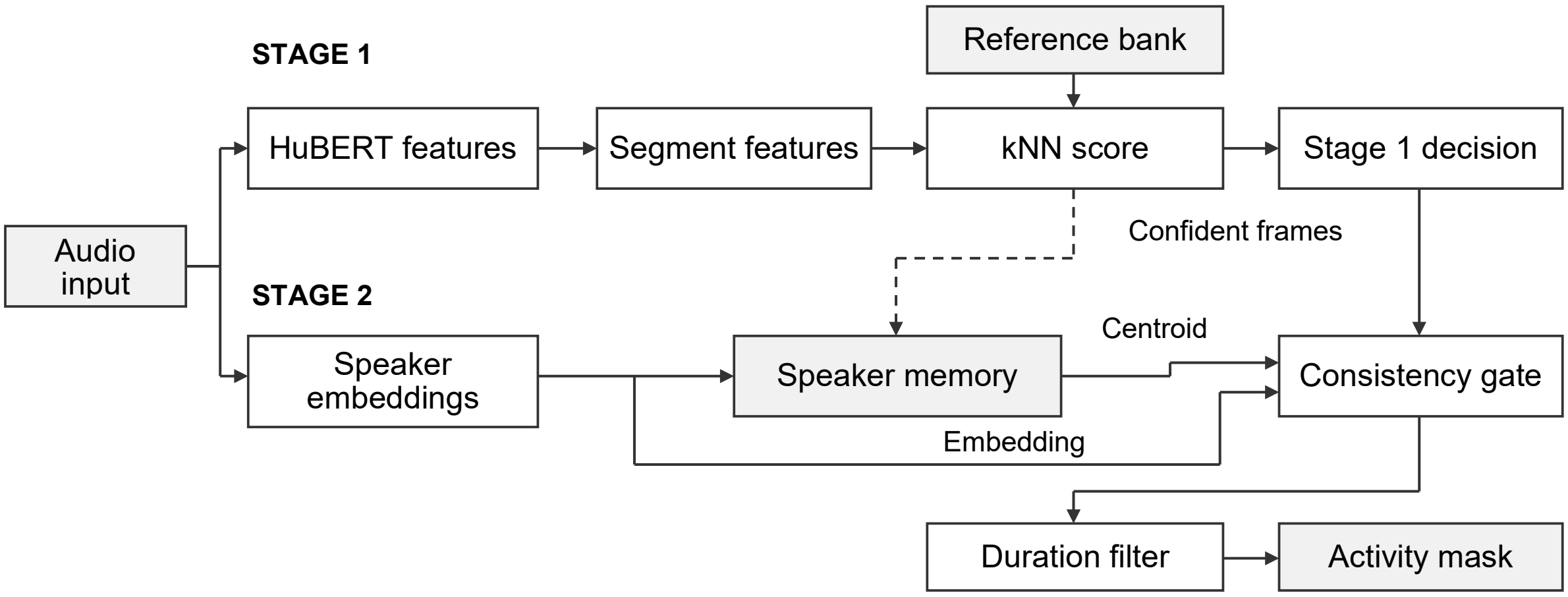

**FIGURE 2. Two-stage foreground one-class VAD; feature transform, reference bank, encoders, and thresholds are fixed before testing.**

TABLE III
VAD PARAMETER SETTINGS AND THE BASIS FOR THEIR SELECTION

| Parameter | Symbol | Value | Rationale and available evidence |
|---|---|---|---|
| Audio and frame | $f_s, \Delta_f$ | 16 kHz; 20 ms | Model: native HuBERT output stride [33]. |
| HuBERT representation | $\ell_H, d_H$ | 7; 768 | Dev.: intermediate layer; width follows the encoder [33]. |
| Encoder block and context | $B_H, L_H, R_H$ | 16; 150; 12 frames | Fixed: streaming geometry; 150 past frames scored higher than 50. |
| PCA dimension and fit cap | $d_P, N_P$ | 64; 50,000 frames | Dev.: compact representation; the cap bounds fitting cost. |
| Pooling context | $L_P, R_P$ | 24; 0 frames | Dev.: 0.5-s window of past frames only; output dimension 128. |
| Reference-bank size | $N_{\text{ref}}$ | ≤100,000 vectors | Dev.: equal draws from the 16 background and SNR cells. |
| Neighbor count | $k_{\text{NN}}$ | 1 | Dev.: distance to the nearest bank vector. |
| Clean-reference activity | $p_E, \Delta_E, E_{\min}$ | 95; 40 dB; -55 dBFS | Dev.: selects the speech frames used for the bank. |
| Minimum label run | $n_{\text{lab}}$ | 1 frame (no removal) | Dev.: no label runs are removed. |
| Pooling speech share | $\eta_P$ | 0.5 | Dev.: at least half of a bank vector's pooling window is labeled speech. |
| Local SNR eligibility | $S_B, S_{\text{cal}}$ | 10 dB; not applied | Dev.: bank and calibration floors selected jointly. |
| Score smoothing | $M_s$ | 3 frames each side | Dev.: a seven-frame mean stabilizes scores; other widths had little effect. |
| Stage 1 quantiles | $q_a, q_u$ | 0.95; 0.90 | Dev.: acceptance and stricter memory admission, selected jointly; SNRs weighted equally. |
| Stage 1 thresholds | $\tau_a, \tau_u$ | 0.153690; 0.136462 | Cal.: positive-score quantiles; fitted before testing and then fixed. |
| Speaker embeddings | $d_e, T_e, H_e$ | 256; 1 s; 0.25 s | Model width [35]; window and hop are fixed design choices. |
| Speaker feature front end | — | 80 mel bins; 25/10 ms; Hamming; pre-emphasis 0.97 | Model-compatible recipe [35]: DC removal, zero dither, per-bin mean subtraction; unit-normalized embeddings. |
| Speaker memory | $T_{\text{mem}}, M_{\min}$ | 240 s; 8 windows | Dev.: retains recent evidence; defers verification until eight windows are stored. |
| Speaker calibration | $q_n, q_c;\ \theta_n, \theta_c$ | 0.20, 0.005; 0.296604, 0.181259 | Dev.: quantile levels selected jointly. Cal.: thresholds fitted on positive calibration data with equal SNR weighting. |
| Minimum output run | $D_{\min}$ | 12 frames (240 ms) | Dev.: suppresses brief detections; may also remove short speech. |

Dev.: value of the configuration with the highest development objective (Section III-D). Fixed: design choice. Model: inherited representation or compatible front end. Cal.: thresholds fitted on positive calibration data.

SNR floor $S_{\text{cal}}$ is not applied in the final configuration. No additional exclusion is applied to centers near speech boundaries.

### *C.* ONLINE SPEAKER CONSISTENCY

Stage 2 extracts unit-normalized, $d_e$ dimensional speaker embeddings with the frozen English VoxCeleb ResNet34 ONNX checkpoint wespeaker_en_voxceleb_resnet34.onnx from WeSpeaker [35]. Its model-compatible log-mel front end is specified in Table III. Embeddings are computed from windows of duration $T_e$, advanced by a hop $H_e$, to provide local speaker evidence at successive times.

Each decision frame is assigned to the nearest speaker-window center, with ties assigned to the earlier center. A window becomes eligible for memory when every decision frame assigned to it satisfies $\bar{d}_t \le \tau_u$. Eligibility is finalized after the last assigned frame is available. This confidence test covers the assigned decision-frame region, which may be shorter than the embedding window. The memory retains eligible windows whose centers lie within the preceding $T_{\text{mem}}$ seconds.

For each stored embedding, neighbors are counted with cosine similarity of at least $\theta_n$. The embedding with the largest neighbor count defines the anchor, and the normalized mean of its neighborhood defines the speaker centroid $c_t$. A Stage 1 candidate passes the consistency gate when

$$g_{\text{spk},t} = 1\left[e_{v(t)}{}^{\mathrm{T}} c_t \ge \theta_c\right] \tag{4}$$

where $e_{v(t)}$ denotes the embedding assigned to frame $t$ and $1[\cdot]$ is the indicator function. The neighborhood threshold $\theta_n$ and gate threshold $\theta_c$ use lower-tail quantile criteria at levels $q_n$ and $q_c$, respectively, on a separate positive calibration set, with equal SNR weighting.

Speaker verification is bypassed by setting $g_{\text{spk},t} = 1$ whenever the memory contains fewer than $M_{\min}$ eligible windows, including when older entries expire. This guard defers rejection until several eligible observations are available. The final activity decision is given by

$$\hat{y}_t = \mathrm{R}_{D_{\min}}\left(1\left[\bar{d}_t \leq \tau_a\right] g_{\text{spk},t}\right) \tag{5}$$

where $R_{D_{\min}}$ removes positive runs shorter than $D_{\min}$ frames. The memory estimates the dominant coherent speaker pattern in recent observations without enrollment. Its effectiveness depends on that pattern corresponding to the designated source. Competing voices, foreground-speaker changes, and erroneous memory updates can affect this correspondence. The gate and duration filter can suppress Stage 1 detections but cannot recover speech rejected by Stage 1; the duration filter can also remove genuine short speech intervals.

### D. PARAMETER SETTINGS

Table III lists the settings used in the reported experiments, their roles, and the basis of their selection. The numerical decision thresholds are fitted from positive calibration data. Settings marked Dev. were selected on the common development panel of Section III-E using a predefined objective: the mean final (Stage 1 + Stage 2) macro-F1 over its 16 background and SNR conditions. The settings were first varied individually and then jointly where they interacted, and the highest-scoring configuration within the 49-frame (0.98-s) look-ahead limit was selected so that the detector's delay stays below 1 s, the step at which the environment analysis receives new audio. Scoring of the test evaluation began only after the configuration was frozen.

A streaming context comparison used the offline training bank: extending past context from 50 to 150 frames increased development macro-F1 at the 98th-percentile operating point without extending future context. The encoder block and future context, and the speaker window and hop, were fixed design choices. The startup count concerns overlapping windows, not independent seconds of speech. Environment-discovery parameters are reported separately in Section IV.

### E. TRAINING AND EVALUATION DATA

Task-specific fitting uses LibriSpeech train-clean-100 dataset [76]. Sixteen utterances from each of the 211 speakers provide speech-plus-noise training mixtures, one for each combination of the four training backgrounds and four SNRs (-5, 0, 5, and 10 dB). A separate group of 20 speakers supplies Stage 1 calibration streams, and another group of 20 supplies the Stage 2 calibration streams. Each calibration set contains every speaker at each of the four SNRs, with the training backgrounds rotated so that each background and SNR combination occurs in the streams. The three speaker groups are disjoint.

The training backgrounds are DKITCHEN, DLIVING, DWASHING, and NFIELD from the DEMAND dataset [77]. Training and calibration use channel 1 and only the first half of each recording. Tests involving these four classes use the second half and assess performance within classes represented during fitting. Stage 1 and Stage 2 thresholds are calibrated using these four DEMAND classes. All parameter selection (Section III-D) uses one common development panel of the LibriSpeech dev-clean speakers in four background conditions: NRIVER from DEMAND; MUSAN sound-bible recordings [78]; streams that switch between these two sources every 20 to 40 s; and ambient recordings from the WHAM! validation split [79]. Each speaker contributes one 8-min base stream per condition, rendered at the four SNRs. STRAFFIC, TBUS, and TCAR are held out from all task-specific fitting and calibration. Every development and training background was compared with every test background by audio fingerprinting and waveform cross-correlation. One of the sound-bible recordings also appears in the FSD50K evaluation set and was removed from development; the test set was not changed. For clips shorter than 3-s, only exact copies could be ruled out; none were found.

The speech test inventory contains utterances from LibriSpeech test-clean speakers and TIMIT test speakers [80]. The TIMIT inventory includes the SA sentences. For each speech corpus and background type, base test streams are added until every background file of the type has been used in full. Every LibriSpeech and TIMIT speaker appears once among the base streams of each type. Each base stream is generated at four SNR levels, resulting in 250.5 h of LibriSpeech test audio and 167.4 h of TIMIT test audio

Background evaluation uses all usable test audio from DEMAND, from the selected FSD50K evaluation categories [81], from the 3,000 WHAM! test noise clips [79], and from six scenes of the TUT Acoustic Scenes 2017 development set (312 ten-second files per scene) [82]. Only ambient recordings are taken from WHAM!; the designated speech sources remain LibriSpeech and TIMIT. Table IV lists the 36 evaluated background types. The FSD50K entries comprise selected labels and grouped label families. Background speech contains clips labeled Crowd, Chatter, Conversation, or Cheering; speech and singing contain the remaining clips with speech or singing labels, so the music type is instrumental only. TUT2017, FSD50K, and WHAM! test recordings do not contribute to task-specific fitting, calibration, or development.

### F. MIXTURE CONSTRUCTION AND REFERENCE LABELS

Each development and test stream concatenates utterances from one designated speaker, retaining entire sentences

TABLE IV

EVALUATED BACKGROUND TYPES AND THEIR RELATION TO TASK-SPECIFIC FITTING

| Corpus | N | Evaluated types | Relation to fitting |
|---|---|---|---|
| DEMAND | 7 | DKITCHEN, DLIVING, DWASHING, NFIELD, STRAFFIC, TBUS, TCAR | Four bank-fitting classes (tests use the second half of each recording); STRAFFIC, TBUS, and TCAR held out from fitting and calibration |
| FSD50K | 22 | Breathing; burping and eructation; cough; crying and sobbing; background speech; instrumental music; knock; laughter; screaming; sigh; sneeze; walking and footsteps; alarms, bells, and signals; animals; domestic and home; non-vocal human sounds; impacts and source-ambiguous sounds; mechanisms, tools, and engines; natural sounds; vehicles; speech; singing | Held out from task-specific fitting and calibration |
| WHAM! | 1 | wham_tt ambient noise | Held out from task-specific fitting and calibration |
| TUT2017 | 6 | bus, cafe/restaurant, city_center, office, park, residential_area | Held out from task-specific fitting and calibration |

N indicates the number of evaluated types. FSD50K entries include grouped label families. Held-out status refers to task-specific fitting and calibration. MUSAN sound-bible recordings, DEMAND NRIVER, and the WHAM! validation split are used only in development.

wherever possible, with an initial gap of 0.8 s. Background audio continues through both utterances and gaps. LibriSpeech test streams last 120 s and development streams 8 min; their inter-utterance gaps, between 0.5 s and at most 6 s, are drawn per stream so that speech occupies 50-60% of the stream where the speaker's utterances allow. The final sentence may end at a natural pause to meet this range. TIMIT streams contain one speaker's ten sentences in random order (about 45 s), separated by gaps of 0.5 to 2.5 s. Background amplitude is set relative to active clean speech at -5, 0, 5, or 10 dB SNR. Each mixture is then scaled, where needed, by its own gain; this preserves the SNR, and the reference labels are computed from the clean speech before any gain, so the four SNR versions of a stream share identical labels. All test streams are evaluated. Negative observations are the source-inactive intervals within these streams.

Primary reference labels are derived from clean speech energy on the 20-ms grid. A frame is positive when its energy exceeds both a threshold 35 dB below the utterance's 99th-percentile energy and an absolute threshold of -65 dBFS. Positive runs shorter than three frames are removed, and internal negative gaps of at most ten frames are filled. Inserted inter-utterance gaps retain their negative labels. These evaluation labels are fixed and separate from the selection label used for fitting (Section III-B). The transition frame, one preceding frame, and two following frames are excluded from scoring. Clean speech references are used offline for label construction and are unavailable to the test detector. An additional TIMIT evaluation derives reference labels from phone annotations, with h#, pau, and epi treated as non-speech. Results from this reference are reported separately because the energy and phone annotations produce different sets of scored frames.

### G. TEMPORAL CONTEXT AND STREAMING INFERENCE

The detector operates with bounded lookahead. The Stage 1 path requires $R_P$ frames for pooling, $M_s$ for smoothing, up to $B_H - 1$ for block alignment, and $R_H$ for encoder context. The duration decision adds up to $D_{\min} - 1$ frames. Their sum, $R_P + M_s + (B_H - 1) + R_H + (D_{\min} - 1)$, is 41 frames for Table III. The steady-state speaker-window path requires approximately 31 future frames; with the duration decision, it sets the overall bound of 42 frames, or 0.84 s on the decision grid. Speaker verification is bypassed during memory startup.

The online memory uses observations available at the delayed decision time. The experiments evaluate bounded-context processing of recorded streams; sustained live throughput and end-to-end device latency require separate measurements.

### H. EVALUATION MEASURES

Overall VAD performance in Table V and the comparisons with both VAE baselines in Table VI are averaged over three fitted seeds. Tables VII and VIII also report three-seed means. All proposed detectors use the 95th-percentile Stage 1 acceptance rule, calibrated before testing and then fixed. Stage 1 denotes the thresholded, smoothed bank score before the final duration filter. The complete system adds both speaker consistency and duration filtering; its comparison with Stage 1 measures their combined effect. Class-specific precision, recall, and F1-score are obtained for each background type from confusion counts pooled over its streams; values for a corpus pair or an SNR condition are means over its background types, so that each type has equal weight. Binary macro metrics average the speech and background values, while accuracy is the proportion of correctly classified frames. Results are reported separately for each speech/background corpus pair and SNR condition.

### I. COMPARISON METHODS

The comparison focuses on one-class task-specific fitting, in which speech-in-noise examples are used without fitting a separate background-only class. ORVAE [43] provides a reconstruction-based comparator for this setting.

TABLE V
VAD PERFORMANCE AND FIRST-STAGE ABLATION BY SPEECH/BACKGROUND CORPUS PAIR, THREE-SEED MEANS (%)

| Background | Speech corpus | Streams | S1 F1 | S1+S2 F1 | Accuracy | Speech recall | Background recall |
|---|---|---|---|---|---|---|---|
| DEMAND | LibriSpeech | 1120 | 89.58 | 91.98 | 92.36 | 94.84 | 88.62 |
| DEMAND | TIMIT | 1120 | 88.87 | 91.21 | 91.79 | 97.64 | 82.99 |
| TUT2017 | LibriSpeech | 960 | 83.72 | 89.47 | 90.11 | 94.19 | 83.98 |
| TUT2017 | TIMIT | 1676 | 85.58 | 89.64 | 90.36 | 96.61 | 80.99 |
| FSD50K | LibriSpeech | 4356 | 83.01 | 89.71 | 90.14 | 91.76 | 87.65 |
| FSD50K | TIMIT | 7672 | 83.32 | 88.57 | 89.24 | 94.51 | 81.34 |
| WHAM! | LibriSpeech | 1080 | 71.36 | 86.87 | 87.58 | 91.94 | 80.98 |
| WHAM! | TIMIT | 2864 | 74.96 | 85.53 | 86.63 | 95.07 | 73.96 |

F1 columns report two-class macro-F1. S1 denotes the first-stage bank detector; S1+S2 adds speaker consistency and short-run filtering. Accuracy and both recalls refer to S1+S2.

The comparison protocol evaluates two reconstruction-based one-class models on the same test streams: an MFCC ORVAE reimplementation based on [43] and a HuBERT residual-VAE adaptation. Because these baselines have no separate calibration step, both are trained using all task-specific fitting recordings of the developed detector. Speech-present training examples are selected from these noisy recordings using activity labels derived from the corresponding clean speech. The selection requires frame energy to exceed both a threshold 30 dB below the utterance's 99th-percentile energy and an absolute threshold of −60 dBFS, a minimum speech run of three frames, and a local SNR of at least 5 dB. These selection settings were kept fixed for the baselines and therefore differ from the final settings in Table III. Consequently, the methods use the same fitting recordings but not the same selected training examples.

The MFCC model uses 80 × 64 trailing patches and a residual convolutional VAE with a 512-dimensional latent space. Its implemented topology contains 2,542,905 parameters, compared with the 4,460,001 reported in the ORVAE paper [43], and is treated as a documented reimplementation. The HuBERT model reconstructs the 128-dimensional pooled representation of the final configuration (Table III) using a residual multilayer-perceptron VAE with hidden width 256 and latent width 32. Both use a full 300-epoch training schedule with three seeds each; the retained checkpoint has the lowest mean reconstruction error on a fixed, stratified sample of development speech frames. The HuBERT model changes both representation and architecture and is therefore an adaptation rather than a feature-only ablation.

Each model is evaluated using the 85th, 90th, 95th, and 97th percentiles of training reconstruction MSE, denoted by T85, T90, T95, and T97. For each seed, a development-selected operating point additionally takes the one of these four percentiles with the highest development objective. An alternative decision uses an OC-SVM [39] on three standardized reconstruction measures: MSE, spectral angle mapper, and one minus structural similarity (SSIM) [83]. The OC-SVM uses an RBF kernel, nu = 0.1, gamma = scale, and a recording-balanced subset of 20,000 training vectors. MFCC SSIM uses a 7 × 7 Gaussian window, whereas HuBERT SSIM uses global population moments across the 128 coordinates with a training-fixed dynamic range. These are explicitly defined adaptations. Each named decision rule is fixed across test conditions; comparison results average the three seed-specific metrics. The comparison results are obtained from the implemented systems and are not taken from the ORVAE paper.

### J. VAD RESULTS AND DISCUSSION

#### 1) OVERALL PERFORMANCE AND THE SPEAKER-CONSISTENCY STAGE

Table V compares the first-stage output with the final detector for each speech/background corpus pair. The final detector achieved three-seed mean macro-F1 values of 86.87%-91.98% with LibriSpeech and 85.53%-91.21% with TIMIT. Adding speaker consistency and short-run filtering improved macro-F1 in all eight pairs, by 2.40-15.51 percentage points for LibriSpeech and 2.34-10.57 points for TIMIT. The largest improvement occurred with WHAM! noise for both speech corpora.

#### 2) COMPARISON WITH THE VAE BASELINES

Tables VI(a) and VI(b) report all four training-percentile settings and the reconstruction-statistic OC-SVM for the two VAE families. The developed detector (S1+S2) achieved higher three-seed mean macro-F1 than every listed VAE setting in all eight corpus pairs. Relative to the reconstruction-statistic OC-SVM, the gains were 20.56-35.97 percentage points for MFCC ORVAE and 2.99-9.69 percentage points for HuBERT VAE. T95 was selected for all three MFCC seeds

TABLE VI(A)
MACRO-F1 FOR THE MFCC ORVAE REIMPLEMENTATION AND THE DETECTOR (%)

| Background | Speech corpus | T85 | T90 | T95 | T97 | OC-SVM | S1 | S1+S2 |
|---|---|---|---|---|---|---|---|---|
| DEMAND | LibriSpeech | 61.58 | 61.86 | 60.25 | 58.35 | 71.42 | 89.58 | 91.98 |
| DEMAND | TIMIT | 51.26 | 54.03 | 58.85 | 62.20 | 63.72 | 88.87 | 91.21 |
| TUT2017 | LibriSpeech | 56.17 | 55.52 | 53.20 | 51.20 | 60.48 | 83.72 | 89.47 |
| TUT2017 | TIMIT | 56.27 | 58.02 | 60.60 | 61.73 | 62.33 | 85.58 | 89.64 |
| FSD50K | LibriSpeech | 63.74 | 68.01 | 72.27 | 73.55 | 68.15 | 83.01 | 89.71 |
| FSD50K | TIMIT | 45.82 | 48.91 | 54.46 | 58.46 | 52.60 | 83.32 | 88.57 |
| WHAM! | LibriSpeech | 46.35 | 45.51 | 43.48 | 41.77 | 56.66 | 71.36 | 86.87 |
| WHAM! | TIMIT | 58.10 | 58.14 | 57.24 | 55.73 | 57.53 | 74.96 | 85.53 |

and T95, T97 and T95 for the HuBERT seeds; these selections were unchanged during testing. Stage 1 alone did not outperform every HuBERT setting; in particular, the OC-SVM performed better with WHAM! for both speech corpora.

The HuBERT reconstruction thresholds produced different balances between speech acceptance and background rejection. Unlike the reconstruction-based baselines, the developed detector also uses a pretrained speaker encoder and an online speaker memory derived from the incoming audio. These architectural differences should be considered when interpreting the improvements; no speaker identity labels or enrollment recordings are supplied during testing.

### 3) SNR DEPENDENCE AND BACKGROUND-SPECIFIC BEHAVIOR

Table VII reports three-seed mean final macro-F1 at each SNR while keeping the two speech corpora separate. Performance generally increased with SNR for TUT, FSD50K, and WHAM!, whereas DEMAND results were comparatively stable. With WHAM!, LibriSpeech macro-F1 increased from 77.86% at -5 dB to 91.93% at 10 dB; the corresponding TIMIT values were 77.52% and 90.64%. All eight corpus pairs exceeded 90% macro-F1 at 10 dB, but low-SNR speech detection remained a limitation.

Table VIII breaks the three-seed mean TUT results down by scene. The verification stage and short-run filter improved macro-F1 in all six scenes for both speech corpora. Their largest gain occurred for cafe/restaurant, a scene containing competing talkers: LibriSpeech improved from 67.90% to 83.55%, and TIMIT from 74.64% to 84.51%. Because these operations remove valid speech as well as false detections, missed speech may enter the intervals selected as background for later environmental analysis.

### 4) REFERENCE SENSITIVITY AND EVALUATION SCOPE

Using TIMIT phone annotations instead of the clean-energy reference produced final macro-F1 values of 91.15%, 89.99%, 88.49%, and 85.50% for DEMAND, TUT, FSD50K, and WHAM!, respectively.

Overall, the results support comparison with a fixed speech-in-noise representation bank followed by recent speaker-consistency verification for the designated-source task evaluated here. Performance remains limited in competing-talker and low-SNR conditions. Within the framework of Fig.

TABLE VI(B)
MACRO-F1 COMPARISON WITH THE HUBERT RESIDUAL-VAE ADAPTATION, ALL TEST STREAMS (%)

| Background | Speech corpus | T85 | T90 | T95 | T97 | Dev. selected | OC-SVM | S1 | S1+S2 |
|---|---|---|---|---|---|---|---|---|---|
| DEMAND | LibriSpeech | 80.12 | 83.25 | 85.92 | 86.36 | 86.12 | 84.71 | 89.58 | 91.98 |
| DEMAND | TIMIT | 80.97 | 84.75 | 87.99 | 88.47 | 88.18 | 85.58 | 88.87 | 91.21 |
| TUT2017 | LibriSpeech | 75.37 | 77.90 | 79.60 | 79.42 | 79.62 | 83.59 | 83.72 | 89.47 |
| TUT2017 | TIMIT | 78.27 | 82.17 | 85.57 | 86.22 | 85.84 | 85.44 | 85.58 | 89.64 |
| FSD50K | LibriSpeech | 74.62 | 77.57 | 79.99 | 80.33 | 80.11 | 80.03 | 83.01 | 89.71 |
| FSD50K | TIMIT | 76.06 | 79.63 | 82.59 | 83.02 | 82.76 | 81.61 | 83.32 | 88.57 |
| WHAM! | LibriSpeech | 67.40 | 68.88 | 68.57 | 66.98 | 68.04 | 79.75 | 71.36 | 86.87 |
| WHAM! | TIMIT | 71.13 | 74.01 | 75.40 | 74.54 | 75.18 | 82.54 | 74.96 | 85.53 |

TABLE VII
FINAL DETECTOR (S1+S2) MACRO-F1 AT DIFFERENT SNRS, THREE-SEED MEANS (%)

| Background | Speech corpus | −5 dB | 0 dB | 5 dB | 10 dB |
|---|---|---|---|---|---|
| DEMAND | LibriSpeech | 91.02 | 92.00 | 92.36 | 92.53 |
| DEMAND | TIMIT | 90.61 | 91.23 | 91.40 | 91.60 |
| TUT2017 | LibriSpeech | 85.09 | 89.37 | 91.23 | 92.04 |
| TUT2017 | TIMIT | 86.11 | 89.66 | 91.13 | 91.57 |
| FSD50K | LibriSpeech | 86.13 | 89.39 | 91.26 | 92.13 |
| FSD50K | TIMIT | 84.92 | 88.16 | 90.17 | 91.12 |
| WHAM! | LibriSpeech | 77.86 | 87.04 | 90.68 | 91.93 |
| WHAM! | TIMIT | 77.52 | 85.06 | 88.87 | 90.64 |

TABLE VIII
MACRO-F1 BY TUT SCENE, THREE-SEED MEANS (%)

| TUT2017 scene | LibriSpeech | | TIMIT | |
|---|---|---|---|---|
| | S1 | S1+S2 | S1 | S1+S2 |
| bus | 89.94 | 92.43 | 90.27 | 91.67 |
| cafe/restaurant | 67.90 | 83.55 | 74.64 | 84.51 |
| city_center | 86.80 | 90.54 | 87.15 | 90.91 |
| office | 79.66 | 87.05 | 83.17 | 87.92 |
| park | 89.26 | 91.94 | 89.11 | 91.46 |
| residential_area | 88.74 | 91.29 | 89.14 | 91.39 |

1, the detector supplies an activity mask for possible subsequent environmental analysis. These VAD scores alone do not establish reliable target-speaker selection or improved clustering or hearing aid performance; those outcomes require a separate evaluation.

## IV. ONLINE CHANGE DETECTION AND RE-IDENTIFICATION OF AUDIO ENVIRONMENTS

Within the components of the developed switching mechanism of Fig. 1, this section addresses the analysis of the background audio that remains after foreground speech has been removed. Three questions must be answered online: whether the environment has changed, whether the current environment has been encountered before, and, if it has, which of the previously discovered environments it is.

The developed method combines a Bayesian online change-point detector, which divides the stream into sound segments, with a segment-level environment memory based on the vigilance principle of the ART2 classifier in [52]. The memory decides whether each detected segment belongs to a known environment or to a new one, and its two decision thresholds are learned from the incoming stream rather than tuned on a dataset. The two learned thresholds are the vigilance $\rho$, the usual tolerance for deciding whether a segment belongs to a previously encountered environment, and the young-segment deviation $e(n)$, an extra allowance while the current segment contains only $n$ embeddings and its mean is still unstable; an open segment of $n$ embeddings is matched with the threshold $\sqrt{\rho^2 + e(n)^2}$. All other parameters, including the alarm probability $\theta_a$ of the detector and the threshold $h$ of the revisit layer, are fixed (see Table IX). A sequential test recognizes a returning environment before the detector confirms the change. Each second is assigned to an environment after a fixed delay, and the assignment is never revised afterward.

### A. CHANGE DETECTION PROBLEM FORMULATION

Let $x_t \in \mathbb{R}^d$, $t = 1,2,\ldots$, denote the audio embedding available at second $t$. The method assigns every second to one of the environments it has discovered so far, which are indexed $1, 2, \ldots$ in the order of their discovery. The assignment $y_t^{(L)} \in \{1, 2, \ldots\}$ of second $t$ is issued at time $t + L$, where the delay $L \geq 0$ is fixed in advance and $L = 0$ gives a live assignment. Two seconds receive the same index if and only if the method judges them to belong to the same environment. The number of environments is unknown and grows as new environments are encountered, and an assignment is final once it has been issued. Because each assignment refers either to a stored environment or to a provisional new one, the output also indicates at every moment whether the current environment has been encountered before. Prototype memory grows only to the adopted cap of $K_{\max}$; issuing a provisional new index does not by itself create a stored environment.

Figure 3 shows the processing chain. The detector monitors the embeddings and, when it confirms a change, reports the second at which the new segment began. The segment that has just ended is then closed: its mean embedding is compared with the environments in memory and is either assigned to the nearest one or stored as a new environment. Between changes, the open segment is assigned every second by comparing its running mean with the memory, using a matching radius that accounts for the larger uncertainty of short segments. A revisit layer watches the open segment for evidence that a known environment has returned and, if so, changes the assignment before the detector responds. The output stage combines these decisions into assignments issued at delay $L$.

### B. AUDIO REPRESENTATION

Audio is converted to mono at 48 kHz and divided into windows of length $T_{\mathrm{w}}$ with hop $T_{\mathrm{h}}$ (Table IX gives the values of all the parameters). Each window is encoded with the pretrained LAION-CLAP audio encoder [65], using the 630k-audioset-best.pt checkpoint with feature fusion disabled, and the 512-dimensional output is normalized to unit Euclidean norm to give $z_t$. The encoder is not adapted. The following causal moving average over $M$ windows reduce fluctuations between consecutive windows:

$$x_t = \frac{\sum_{j=0}^{M-1} z_{t-j}}{\left\|\sum_{j=0}^{M-1} z_{t-j}\right\|_2} \tag{6}$$

$x_t$ depends only on the current and the $M-1$ preceding windows. The embedding dimension is $d = 512$. At startup, the average uses the windows available so far; the implementation includes a small normalization safeguard for a zero norm.

### C. ONLINE CHANGE DETECTION

Changes are detected with Bayesian online change-point detection (BOCPD) [72]. BOCPD maintains the posterior distribution of the run length $r_t$, the number of embeddings since the most recent change. With a constant hazard $H$, the run length grows by one with probability $1-H$ or returns to zero with probability $H$ at each step, and the joint distribution is updated recursively as follows:

$$p(r_t, x_{1:t}) = \sum_{r_{t-1}} p(r_t | r_{t-1}) \times \pi(x_t | r_{t-1}) p(r_{t-1}, x_{1:t-1}) \tag{7}$$

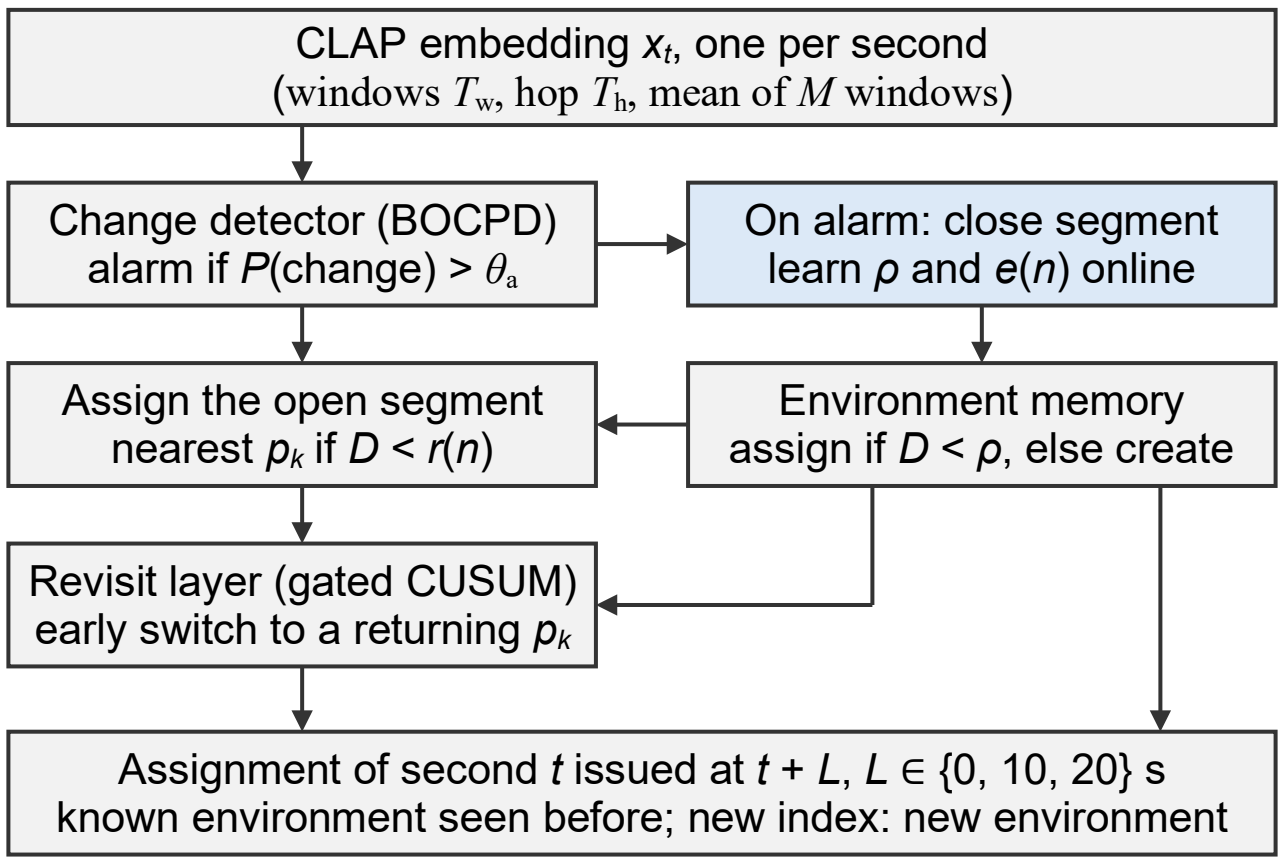

**FIGURE 3. Developed online environment discovery. A provisional new index becomes a stored environment only when a completed sound segment is assigned to it. The blue shaded block learns the vigilance $\rho$ and the young-segment deviation $e(n)$ from the stream; $L$ denotes the output delay.**

where $\pi(x_t | r_{t-1})$ is the predictive density of $x_t$ given the embeddings of the current run. Each dimension of $x_t$ is modeled as Gaussian with unknown mean and precision under a conjugate Normal-Gamma prior with parameters $\mu_0$, $\kappa_0$, $\alpha_0$ and $\beta_0$. The dimensions are treated as independent, so $\pi$ is a product of 512 Student-$t$ densities. The hazard $H$ corresponds to an expected run length of $1/H$ seconds. For efficiency, following [72], run lengths beyond the point at which the cumulative posterior reaches $1-\varepsilon_{\mathrm{R}}$ are discarded. Additionally, at most $R_{\max}$ run lengths are retained.

BOCPD does not specify when a change should be declared. Adapting the alert convention in [75], an alarm is raised at time $t$ when the posterior probability of a change since the previous alarm, at time $a$, exceeds $\theta_{\mathrm{a}}$:

$$\sum_{r=0}^{t-a} p(r_t = r | x_{1:t}) > \theta_{\mathrm{a}}. \tag{8}$$

The change location is then $c = t - r^* + 1$, where $r^*$ is the most probable run length between 1 and $t-a$, so that the new segment starts at second $c$. An alarm whose change location lies within $T_{\mathrm{m}}$ of the previous change location is ignored. The detector thus reports events $(t, c)$: the change becomes known at time $t$ and is placed at $c \le t$. On the development streams, changes are confirmed about 10 s after they occur, which motivates the revisit layer and the delayed outputs described below.

### D. ENVIRONMENT MEMORY

The memory stores one prototype $p_k \in \mathbb{R}^d$ for each discovered environment $k = 1, \ldots, K$, together with the number $N_k$ of segments assigned to it. Adapting the distance rule of the ART2 classifier in [52], dissimilarity is measured with the mean absolute difference

$$D(u, v) = \frac{1}{d} \sum_{i=1}^{d} |u_i - v_i| \tag{9}$$

and a vigilance threshold $\rho$ decides whether an input is close enough to its nearest prototype to be assigned to it. Unlike [52], which classifies every frame, the memory is updated only when the detector closes a segment. For a closed segment $[s, c)$, with start $s$ and change location $c$, the input is the average of its $c-s$ embeddings, $m(s,c) = (x_s + x_{s+1} + \cdots + x_{c-1})/(c-s)$. With $k^* = \operatorname{argmin} D(m(s,c), p_k)$, the segment is assigned to environment $k^*$ if $D(m(s,c), p_{k^*}) < \rho$, and the prototype is updated as follows:

$$p_{k^*} \leftarrow \frac{N_{k^*} p_{k^*} + m(s,c)}{N_{k^*} + 1}, \qquad N_{k^*} \leftarrow N_{k^*} + 1. \tag{10}$$

If the distance is strictly greater than the threshold and capacity is available, a new environment $K+1$ is created with $p_{K+1} = m(s,c)$ and $N_{K+1} = 1$; the first closed segment always creates the first environment. All seconds of the closed segment are assigned to the environment decided at closing.

The adopted design cap, taken from the released implementation accompanying [52], is $K_{\max}$ environments; when it is full, a segment that matches no prototype is assigned to the same environment as the preceding segment. Between changes, the open segment $[s_t, t]$ is assigned every second by applying the same comparison to its running mean, without updating the memory. If no prototype is close enough, the open segment receives the provisional index $K+1$, the index that a new environment would obtain if the segment were closed at that moment.

### E. STREAM-LEARNED VIGILANCE

The vigilance is intended to distinguish matching from nonmatching segment prototypes; its calibration uses detected segments without access to class labels. Its useful value depends on the embedding, the distance measure and the audio characteristics, so a value tuned on one dataset need not transfer to another. The developed method therefore learns $\rho$ from the stream, using two statistics that become available whenever a segment is closed. For a closed segment $[s, c)$ with at least $N_W$ windows, the within-segment statistic compares the means of its two halves,

$$W = D\big(m(s, h_{\mathrm{mid}}), m(h_{\mathrm{mid}}, c)\big), \quad h_{\mathrm{mid}} = s + \lfloor c - s/2 \rfloor, \tag{11}$$

and measures the heterogeneity of two parts of a detected segment. For a closed segment with at least $N_B$ windows, the between-segment statistic compares its mean with the mean $m'$ of the most recent earlier closed segment that also contained at least $N_B$ windows,

$$B = D(m(s, c), m'), \tag{12}$$

and is used as a proxy for separation between environments. The vigilance is placed halfway between the typical values of the two statistics,

$$\rho = \frac{1}{2}[\mathrm{med}(\mathcal{W}) + \mathrm{med}(\mathcal{B})], \tag{13}$$

where $W$ and $B$ collect all values observed thus far in the stream, and $\mathrm{med}(\mathcal{W}) = 0$ while $W$ is empty. At each closing, $W$ and $B$ are computed first, $\rho$ is updated, and the segment is then assigned with the updated value. Before the first value of $B$ exists, $\rho = 0$, so the first closed segment creates the first environment. The use of medians is intended to make $\rho$ robust to occasional false alarms, which can produce small values of $B$, and to missed changes, which can produce large values of $W$.

### F. YOUNG-SEGMENT MATCHING RADIUS

The running mean of a young open segment is a noisy estimate of the mean of its environment: after $n$ windows, its distance to the correct prototype includes sampling variation that can change as $n$ grows. Applying the closing threshold $\rho$ to young segments would reject returning environments and produce provisional new indices that later prove incorrect. The size of this deviation is also learned from the stream; $e(n)$ is the second of the two learned thresholds. When a segment $[s, c)$ is closed, the deviation $\varepsilon_{s,c}(n) = D\big(m(s, s+n), m(s, c)\big)$ of the mean of its first $n$ windows from the mean of the whole segment is recorded for every $n = 1, \dots, n_{\max}$ with $c - s \geq 2n$, and

$$e(n) = \begin{cases} \mathrm{med}\, \varepsilon_{s,c}(n), & n \leq n_{\max}, \\ e(n_{\max})\sqrt{n_{\max}/n}, & n > n_{\max}, \end{cases} \tag{14}$$

where the median is taken over the closed segments for which $\varepsilon_{s,c}(n)$ has been recorded, and $e(n) = 0$ until a value exists. The extension beyond $n_{\max}$ windows follow the $1/\sqrt{n}$ scaling as a chosen extrapolation. An open segment of $n$ windows is assigned to the environment of its nearest prototype $k^*$ if

$$D(m(s_t, t+1), p_{k^*}) < r(n) = \sqrt{\rho^2 + e(n)^2} \tag{15}$$

and receives the provisional new index otherwise. The radius combines the spread of an environment, represented by $\rho$, with the sampling deviation of a young mean, and beyond $n_{\max}$ windows the prescribed extrapolation approaches $\rho$ as the segment grows. Closing decisions continue to use $\rho$, so the radius affects only the assignments of open segments and never the memory.

### G. REVISIT LAYER

When a known environment returns, the detector confirms the change only after several seconds, and until then the open segment, still dominated by the previous environment, remains assigned to it. The revisit layer shortens this period with a CUSUM test [71] for each known environment. Let $u_k = p_k/\|p_k\|_2$ be the unit-norm prototype of environment $k$, and let $q_t$ be the unit-norm mean of the open segment without its most recent $G$ seconds, that is, of $x_{s_t}, \dots, x_{t-G}$. For each known environment other than the current assignment, the evidence that $x_t$ comes from $k$ rather than from the open segment is

$$g_k(t) = {x_t}^T u_k - {x_t}^T q_t, \tag{16}$$

and it is accumulated with the slack $\delta$ as

$$S_k(t) = \max\{0, S_k(t-1) + g_k(t) - \delta\}. \tag{17}$$

The onset time $o_k$ marks the beginning of the current interval during which $S_k$ remains positive. The layer changes the assignment to the environment with the largest statistic when $S_k(t) > h$, after at least $n_{\min}$ seconds of positive evidence, and the mean of the embeddings since the onset agrees with the prototype:

$$1 - {v_k}^T(t) u_k \leq \gamma, \tag{18}$$

where $v_k(t)$ is the unit-norm mean of $x_{o_k}, \dots, x_t$ and the gate $\gamma$ is a cosine distance. The gate is intended to reduce switches to an environment whose prototype is merely closer than the open segment to a few transient embeddings. After a switch, a reverse statistic with evidence ${x_t}^T q_t - {x_t}^T u_k$ is accumulated in the same way, and the layer returns to the base assignment when this statistic exceeds $h$. The layer is reset whenever the assignment of the open segment changes, new evidence accumulation is enabled only when the open segment is at least $A_{\min}$ seconds old, and it never creates environments. A boundary that leaves the base identity unchanged need not reset an already active revisit assignment.

### H. ASSIGNMENTS ISSUED AT A FIXED DELAY

At every second $t$, the live assignment $y_t^{(0)}$ is the assignment of the open segment, replaced by the revisit assignment when that layer is active. For a delay $L > 0$, the assignment of second $\tau$ is issued at time $\tau + L$ using information available then. The base delayed assignment uses the closed-segment identity if its segment has closed, and otherwise the current open-segment identity. When the delayed assignment for second $\tau$ is issued, an active revisit assignment replaces the base assignment if $\tau \geq o_k$; even if the segment containing that second has already closed.

The delay trades responsiveness for accuracy. With $L = 10$ s, most changes have been confirmed by the time the affected seconds are assigned, while the delay remains short compared with the time usually spent in an environment, so the delayed assignments can still drive the selection of environment-specific settings, such as the personalized gain profiles of [14]. Fig. 4 shows the three outputs on a development stream. Live assignments usually follow a change only after the detector alarm or a switch of the revisit layer, whereas assignments issued after 10 or 20 s align more closely with the true changes. In our evaluation, processing stops at the end of each recording. The remaining delayed assignments are filled using the base assignments available at the last update, before the final segment is closed, without applying the revisit layer. The reported all-window delayed scores include this endpoint fill.

### I. ALGORITHM, PARAMETERS AND COMPUTATIONAL COST

Algorithm 1 summarizes one step of the method. Every decision uses only embeddings up to the current second, detector events already raised and the earlier decisions of the method. Table IX lists every parameter with its value, the basis or justification for selection of the values. The basis is prior work, a design argument, selection on the development streams (Section IV-J3), learning from the stream, or the intended application. The eight values with the development basis were selected once on the development streams and were kept fixed for every test corpus; their effect is examined in Section IV-M.

**Algorithm 1.** Online environment discovery at second $t$

```
update the run-length posterior with x_t (7)
if an alarm (8) places a change at c then
    compute m = m(s, c), append W (11) and B (12), update ρ (13)
    assign m to k* and update its prototype (10), or create environment
    K + 1; assign seconds s, ..., c − 1 to it
    record ε_{s,c}(n), update e(n) (14); set s ← c
end if
assign the open segment using the radius r(n), n = t − s + 1 (15)
if n ≥ A_min then update the revisit layer (16)-(18)
issue the live assignment of t and the assignment of t − L for each delay
L > 0 (Section IV-H)
```

Per second, the BOCPD update requires $O(R_t d)$ operations, where $R_t \leq R_{\max}$ is the number of retained run lengths; matching against the memory and updating the revisit layer require $O(Kd)$ operations each, with $K \leq K_{\max}$. Computing a segment mean from prefix sums requires $O(d)$ operations and computing new prefix deviations at a closing requires $O(n_{\max} d)$ operations. Exact median evaluation adds computation that grows with the stored calibration histories. Detector state occupies $O(R_t d)$ memory and prototypes occupy $O(Kd)$. In the evaluated implementation, cumulative statistics are retained for the full stream, so their storage grows proportionally to the number of processed windows times the embedding dimension. The sets $W$ and $B$ and the deviations for $e(n)$ also grow with the number of completed segments. The implementation is causal but does not have a total-memory bound independent of duration.

### J. EVALUATION PROTOCOL

#### 1) DATA

The main evaluation used the official test partitions of two corpora. TAU Urban Acoustic Scenes 2019 [84] contains 10-s clips grouped by recording location. The reported TAU results use bus, park, shopping mall and street traffic recordings from the official fold-1 evaluation split. CochlScene [85] contains 10-s excerpts from recordings of about 1 min collected with smartphones. The reported CochlScene results use car, kitchen, restaurant and street recordings from the official test partition.

#### 2) TEST STREAMS

Two families of test streams were constructed for each corpus. In recording-continuous (RC) streams, each block contains audio of one scene and keeps the original order of the windows within each 10-s clip, but the clips of the scene are taken in

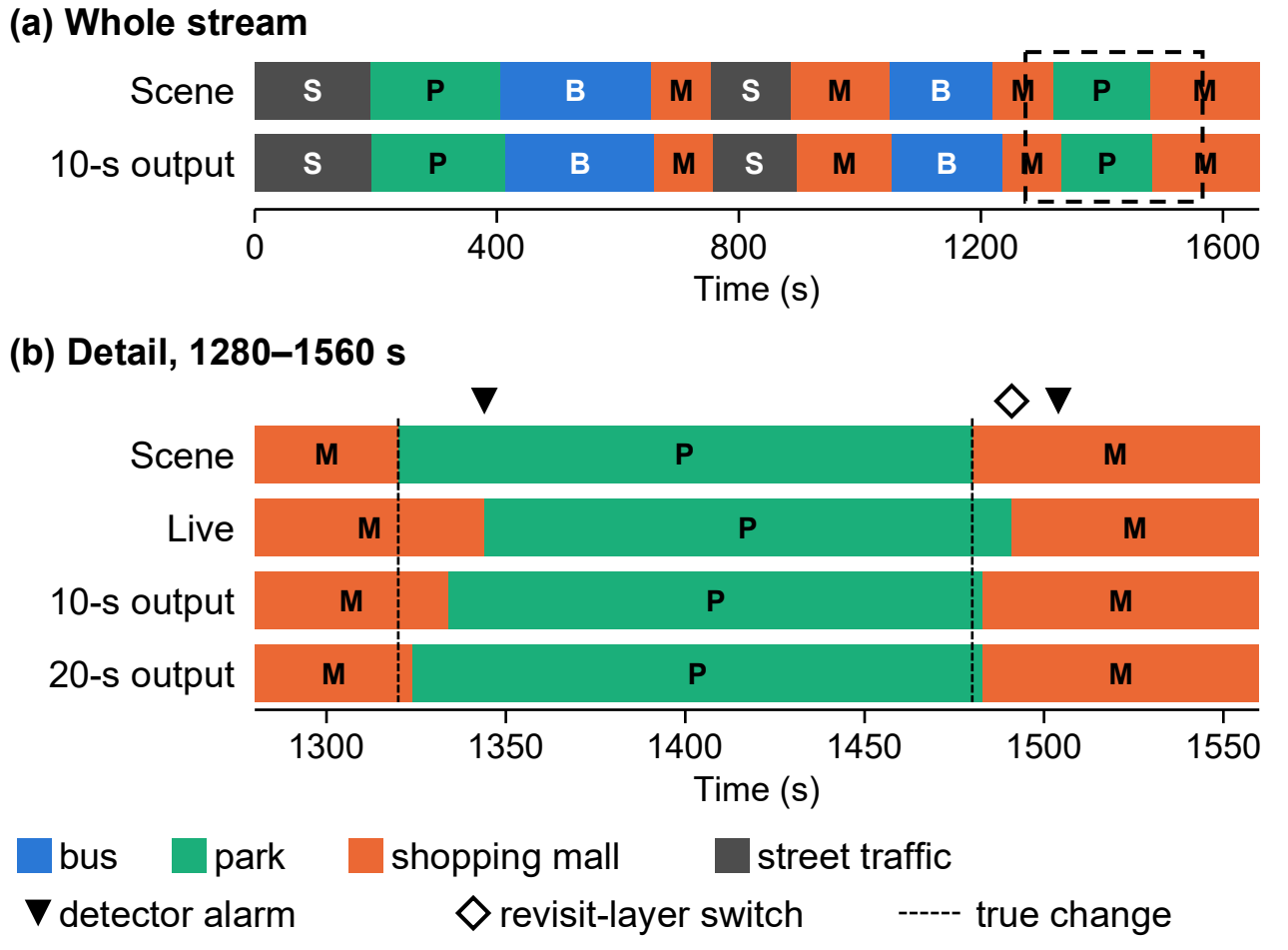

**FIGURE 4. Example operation of the developed method on a TAU stream (recording-continuous, random duration). (a) Whole stream: scenes and assignments issued at a 10-s delay. (b) Detail of 1280-1560 s: scenes and assignments issued live, at 10 s and at 20 s. Each discovered environment is colored as the scene to which it is matched for display; letters give the scene (B, bus; P, park; M, shopping mall; S, street traffic).**

TABLE IX

PARAMETERS OF THE DEVELOPED METHOD

| Parameter | Symbol | Value | Selection basis or justification |
|---|---|---|---|
| *Representation (Section IV-B)* | | | |
| Audio encoder | – | LAION-CLAP; 48 kHz; $d = 512$ | Prior work: pretrained general-audio encoder [65], used without adaptation; the input rate and the dimension are those of the released model. |
| Window, hop, smoothing | $T_{\mathrm{w}}, T_{\mathrm{h}}, M$ | 2 s, 1 s, 3 windows | Design: $T_{\mathrm{h}}$ gives one assignment per second; $T_{\mathrm{w}} = 2T_{\mathrm{h}}$ gives 50% overlap at low latency; $M$ averages 4 s of audio against brief foreground sounds for about 1 s of delay. Best of 1, 2, 3, 5 and 7 windows on the development streams. |
| *Change detector (Section IV-C)* | | | |
| Hazard rate | $H$ | 1/300 | Development: a prior expectation of 5 min between changes; results unchanged from 1/100 to 1/10,000. |
| Normal–Gamma prior | $\mu_0, \kappa_0, \alpha_0, \beta_0$ | 0, 0.01, 1, 0.01 | Design ($\mu_0$): neutral for unit-norm embeddings and weighted by only $\kappa_0$ observations; the first embedding instead left the results unchanged. Development ($\kappa_0$, $\alpha_0$, $\beta_0$): weak priors ($\kappa_0 = 0.01$ and $2\alpha_0 = 2$ pseudo-observations); $\beta_0$ is the sensitive value, and $J$ peaks at $\beta_0/\alpha_0 = 0.0075 - 0.01$ in every development corpus. |
| Alarm probability | $\theta_{\mathrm{a}}$ | 0.95 | Prior work: alert convention of [75]; results unchanged from 0.8 to 0.99. |
| Merge interval | $T_{\mathrm{m}}$ | 5 s | Design: longer than the 4-s span of a smoothed embedding, so that one change gives one boundary; identical results from 0 to 10 s. |
| Run-length truncation | $\varepsilon_{\mathrm{R}}, R_{\max}$ | $10^{-4}$, 2000 | Prior work: negligible run lengths are pruned as in [72]. Design: a computational cap, eight times the longest detected development segment, never reached. $\varepsilon_{\mathrm{R}} = 10^{-6}$ or $R_{\max} = 500$ gave identical results. |
| *Environment memory and learned thresholds (Sections IV-D to IV-F)* | | | |
| Distance | $D$ | Mean absolute difference, (9) | Prior work: distance rule of [52]. The Euclidean distance gave the same ARI and the cosine distance a lower one. |
| Memory capacity | $K_{\max}$ | 10 | Prior work: released implementation of [52]. Never reached on the development streams (at most five environments stored); identical results without a limit. |
| Vigilance | $\rho$ | Learned | Learned from each stream by (13); no selected value. |
| Weight in (13) | – | 1/2 | Design: halfway between the typical same-environment and different-environment distances. Best of 0.3 to 0.7 on the development streams. |
| Minimum length for $W$, $B$ | $N_W, N_B$ | 40 s, 20 s | Design: every mean entering $W$ or $B$ covers at least 20 s, five times the smoothing span; 99% of the detected development segments qualify for $W$. Count and $J$ unchanged when halved or doubled. |
| Young-segment deviation | $e(n)$ | Learned | Learned from each stream by (14); no selected value. |
| Longest measured prefix | $n_{\max}$ | 60 windows | Design: the longest prefix that most segments can supply; with $c - s \geq 2n$ it needs segments of 120 s, met by 89% of the detected development segments (240 s: 2%). 30 or 120 left the ARI unchanged. |
| Prefix limit; extrapolation exponent | – | $c - s \geq 2n$; 1/2 | Design: a prefix of at most half of the segment does not dominate the mean it is compared with; the standard error of a mean falls as $n^{-1/2}$. Factors 1.5 and 3 and exponents 0.25 and 1 had no effect. |
| *Revisit layer (Section IV-G)* | | | |
| CUSUM slack, threshold | $\delta, h$ | 0.20, 0.1 | Development: selected together with $\gamma$ and $n_{\min}$ (Section IV-J3). RC ARI within 0.003 over $\delta = 0.02 - 0.20$ and $h = 0.05 - 0.3$; $h \geq 2$ effectively disables the layer. |
| Gate (cosine distance) | $\gamma$ | 0.075 (22.3°) | Development: same selection. In a development check of the layer without the gate, wrong switches matched the chosen environment at 23° or more and correct ones at a median of 16°; the median angle between a 2-s stretch of an environment and its mean is 22.2°. Larger $\gamma$ trades false switches for ARI (Fig. 5). |
| Minimum evidence run | $n_{\min}$ | 1 s | Development: same selection. RC ARI 0.946-0.947 over 1-5 s. |
| Excluded recent window | $G$ | 5 s | Design: longer than the 4-s span of a smoothed embedding, so that $q_t$ excludes embeddings that may already contain the new environment; no effect over 2-10 s. |
| Minimum segment age | $A_{\min}$ | 20 s | Design: the same 20-s minimum as for the means entering $W$ and $B$; count and $J$ unchanged over 10-40 s. |
| *Output (Section IV-H)* | | | |
| Output delay | $L$ | 0, 10, 20 s | Application: latency against accuracy; all three outputs are reported (Section IV-H). |

Prior work: taken from the cited work. Design: fixed by the stated argument. Development: selected once on the development streams of TUT, SINS and HEAR-DS and kept fixed for every test corpus. Learned: the two thresholds $\rho$ and $e(n)$, computed from each stream during operation; all other values are fixed. Application: set by the intended use. Effects: one value varied at a time with the developed method on the TUT, SINS and HEAR-DS development streams, as RC ARI at the 10-s output and the objective $J$ of Section IV-J3 (Section IV-M).

random order, so consecutive clips usually come from different recordings. In non-contiguous (NC) streams, the windows of a block are drawn in random order from the test windows of its scene, which removes the continuity within recordings. Each family contains five class-wise streams, with two 250-s blocks per scene arranged so that consecutive blocks differ, and 25 random-duration streams with blocks of 80-250 s. For TAU, the 30 streams of each family contain about 21 h of audio and more than 400 environment changes, of which 320 (RC) and 311 (NC) return to an environment already visited in the same stream. The streams reuse the test recordings in different orders and durations, so they are not independent samples of recordings. The TAU test streams draw on 53 recordings. The CochlScene test streams follow the same construction, with 30 RC and 30 NC streams drawn from 380 test recordings of 142 participants (94 to 96 recordings per scene).

### 3) DEVELOPMENT DATA AND PARAMETER SELECTION

For TUT Acoustic Scenes 2017 [82], 20% of the recordings of each class, were selected randomly with a fixed seed, giving 10, 13 and 16 development recordings for TUT-3, TUT-4 and TUT-5, respectively. For SINS [86], three sessions per class were selected with a fixed seed, two of at least 200 s and one shorter, preserving long sessions for the test blocks. For HEAR-DS [87], which contains only three or four recording situations per class, the lowest-numbered situation was used for development. Recordings or sessions used for development were excluded from every test pool across the class sets. All methods used the same splits. Smaller development shares were used instead of the official cross-validation folds of TUT and SINS to retain larger recording pools for test-stream construction. For each class set, eight RC and eight NC development streams were constructed as in Section IV-J2, comprising five class-wise and three random-duration streams in each family. This produced 96 development streams for TUT, SINS and HEAR-DS.

The eight numerical parameters identified as development-selected in Table IX were selected using only the TUT, SINS and HEAR-DS development streams. The four detector values were selected with the detector-and-memory configuration of Section IV-L, which uses the same detector, as one setting for all class sets. The selection objective, based on the adjusted Rand index (ARI) [88], was $J = ½[\mathrm{mean}(\mathrm{ARI}) + \min(\mathrm{ARI})] + ½P_K$, where $\mathrm{mean}(\mathrm{ARI})$ and $\min(\mathrm{ARI})$ are the mean and the lowest ARI over the RC and NC streams of a class set and $P_K$ is the share of these streams with the correct number of environments; $J$ was computed with end-of-stream assignments and averaged over class sets and then over corpora. The four revisit-layer values were selected from a grid of 225 settings, with the layer applied to the fixed-vigilance detector-and-memory configuration of Section IV-L. The selected setting maximized live ARI without increasing live false switches relative to the same configuration without the layer. The resulting method was frozen before the reported TAU test evaluation and applied unchanged to CochlScene. No settings were reselected for the 10-s output.

### 4) OUTPUTS AND MEASURES

The live output and the 10-s (main) and 20-s outputs are evaluated on all windows, with the delayed-output endpoint convention stated in Section IV-H. For test scoring, the dataset scene annotations provide the reference labels. For each stream, the ARI and the normalized mutual information (NMI) [89], with arithmetic-mean normalization, compare the issued assignments with the scene annotations, and the accuracy (ACC) is computed after an optimal one-to-one matching of discovered environments to scenes [90]. Exact $K$ indicates whether the number of distinct environments in the issued assignments equals the true number of environments, and the percentage of streams with exact $K$ is reported. Exact $K$ is the primary count measure because, in the intended hearing aid application, every issued index selects a gain profile, so an extra index means an unnecessary profile. A merge of two scenes can, however, be offset by an extra index and leave exact $K$ correct. A scene is therefore also counted as absorbed when the environment to which most of its seconds are assigned is matched to another scene in the one-to-one matching, and the identity-aware count is the percentage of streams with exact $K$ and no absorbed scene. False switches per hour count the changes of assignment that cannot be matched to a true change: each true change is matched to at most one change of assignment within ±30 s, and the remaining changes are counted. Memory $K$ indicates whether the memory holds the true number of environments at the end of the stream.

Recognition of seen and new environments is evaluated per visit, a visit being a maximal run of one environment; the first visit of each stream is excluded. A revisit is recognized if most seconds of the visit are assigned to the same discovered environment as most seconds of the previous visit to that audio scene, and a new environment is flagged if the environment to which most seconds of its first visit are assigned does not occur among the assignments issued before that visit. The time to the correct assignment is the time from the start of a visit until the first issued assignment that corresponds to its scene under the one-to-one matching, including the output delay. Because a visit counts as recognized when most of its seconds are assigned correctly, the 95th percentile and the maximum of this time over all visits are also reported. Clustering measures and false-switch rates are averaged over the 30 streams of a family. Recognition percentages pool audio scene visits across streams. Timing summaries use the stated mean of per-stream medians or pooled-visit quantiles and maxima. RC and NC streams are reported separately.

### 5) TEST STREAMS OF TUT, SINS AND HEAR-DS

The test splits of the three development corpora were also processed: TUT Acoustic Scenes 2017 [82] with three nested class sets (TUT-3: bus, office and residential area; TUT-4: TUT-3 and café/restaurant; TUT-5: TUT-4 and metro station), SINS [86] with cooking, vacuum cleaner and watching TV (SINS-3), and HEAR-DS [87] with cocktail party, in traffic, in vehicle, quiet indoors and wind turbulence (sep5) and with the same classes except quiet indoors (noise4). Each class set has six RC and six NC test streams, one class-wise and five of random duration, built from recordings that no development stream used. These streams sample the test audio rather than using all of it: the TUT streams use 72-100% of the test windows of each class and the HEAR-DS streams 30-58%, and the SINS RC streams play each session from its beginning and leave four of the 22 test sessions unused. These streams were examined while the method was developed, so their results show its behavior on further corpora.

### K. OUTCOME

#### 1) OVERALL PERFORMANCE

Table X summarizes the results. At the main 10-s delay, the RC streams reached an ARI of 0.912; the number of environments was correct in 24 of the 30 streams (80.0%), and false switches occurred at an average rate of 1.0 per hour (median 0). The NC streams reached an ARI of 0.962, with the correct number of environments in all 30 streams and 0.05 false switches per hour. A 20-s delay raised the RC ARI to 0.952, with the correct number in 29 streams (96.7%). The live assignments were less accurate (RC ARI 0.806, NC 0.917) because the seconds that follow a change remain assigned to the previous environment until the change is recognized.

At the end of every stream, in both families, the memory held exactly four environments, the true number. Every count error was therefore in the issued assignments, not in the memory. In each of the six RC streams with an incorrect count at 10 s (all of them random-duration streams), the error was one extra index: a segment that could not yet be matched to a stored environment when its first assignments were issued received a provisional new index, and when it was closed, it was assigned to a known environment, so no new environment was stored. Because issued assignments are never revised, that provisional index remained in the output.

#### 2) PREVIOUSLY SEEN AND NEW ENVIRONMENTS

Table XI shows that the method recognized nearly every return to a known environment: 319 of 320 revisits in the RC streams and all 311 in the NC streams at the 10-s output. New environments were flagged as new in 88 of 90 cases (RC) and in all 90 (NC). With live assignments, the mean of the per-stream median times to the correct assignment was 12.6 s in the RC streams (11.9 s for revisits and 17.2 s for new environments) and 4.0 s in the NC streams, and every visit received the correct assignment before it ended. At 10 s, the 95th percentile of the time to the correct assignment over all visits was 32 s in the RC streams and 20 s in the NC streams. New environments took longer than revisits, as expected: the revisit layer can recognize a returning environment before the change is confirmed, whereas a new environment is recognized as new only once its segment differs from every stored environment by more than the matching radius.

TABLE X
PERFORMANCE OF THE DEVELOPED METHOD ON THE TAU TEST STREAMS

| Family | Output | ARI | NMI | ACC | Exact K (%) | False sw./h |
|---|---|---|---|---|---|---|
| RC | Live | 0.806 | 0.769 | 0.922 | 63.3 | 1.98 |
| RC | 10 s | 0.912 | 0.884 | 0.966 | 80.0 | 1.02 |
| RC | 20 s | 0.952 | 0.931 | 0.982 | 96.7 | 0.36 |
| NC | Live | 0.917 | 0.885 | 0.968 | 100 | 0.28 |
| NC | 10 s | 0.962 | 0.942 | 0.986 | 100 | 0.05 |
| NC | 20 s | 0.975 | 0.958 | 0.990 | 100 | 0.00 |

Mean over the 30 streams of each family. Exact K: percentage of streams in which the number of environments in the issued assignments equals the true number (four). False sw./h: changes of assignment not matched to a true change, per hour. Memory K was correct in all 60 streams.

#### 3) STREAM TYPES AND SCENES

Table XII separates the two stream types. Class-wise streams, with two long blocks per scene, were easier than random-duration streams, whose blocks can be as short as 80 s and which contain about twice as many changes. At 10 s, the RC ARI was 0.949 for class-wise and 0.904 for random-duration streams, and all six count errors occurred in random-duration streams. The recall of each scene after the one-to-one matching ranged at 10 s from 0.951 (park) to 0.974 (bus and shopping mall) in the RC streams and from 0.979 (park) to 0.992 (bus) in the NC streams, and no scene was absorbed by another in any stream. Park, a member of the most confusable pair in the class-selection analysis, had the lowest recall in both families.

#### 4) CONTRIBUTION OF THE COMPONENTS

Table XIII removes the components one at a time and in combination. Without the revisit layer, the RC ARI fell from 0.912 to 0.861 at 10 s and from 0.806 to 0.744 for live assignments, and false switches increased in both families, while the number of environments was unaffected. Without the young-segment radius, the ARI was nearly unchanged, but the issued count was correct in only 40.0% of the RC streams and false switches more than doubled, because young segments of returning environments received provisional new indices more often. Replacing the learned vigilance with the value selected on the development streams gave the same ARI and a similar count accuracy (83.3% vs. 80.0%, a difference of one stream); the vigilance learned from the stream thus matched a tuned threshold without requiring one. The detector and memory alone, that is, the classifier of [52] applied to detected segments, reached an RC ARI of 0.859 and the correct count in 33.3% of the streams. With the revisit layer but without the two learned thresholds, the RC ARI was 0.910, and the count was again correct in only 33.3% of the streams. With the vigilance selected on the RC development streams only, the count was correct in 93.3% of the RC streams, but the looser vigilance merged scenes.

#### 5) COCHLSCENE

Table XIV gives the results on CochlScene. The NC streams were handled as well as on TAU: at 10 s, the ARI was 0.969,

TABLE XI
RECOGNITION OF SEEN AND NEW ENVIRONMENTS AND TIME TO THE CORRECT ASSIGNMENT

| Family | Output | Revisits recog. (%) | New flagged (%) | Time to correct assignment (s) | | |
|---|---|---|---|---|---|---|
| | | | | All | Revisit | New |
| RC | Live | 99.7 | 97.8 | 12.6 | 11.9 | 17.2 |
| RC | 10 s | 99.7 | 97.8 | 13.4 | 13.1 | 17.7 |
| RC | 20 s | 99.7 | 95.6 | 21.2 | 21.3 | 22.4 |
| NC | Live | 99.7 | 98.9 | 4.0 | 3.4 | 13.8 |
| NC | 10 s | 100 | 100 | 11.7 | 11.6 | 14.9 |
| NC | 20 s | 100 | 100 | 21.2 | 21.2 | 21.7 |

Times are per-stream medians averaged over the 30 streams and include the output delay.

TABLE XII
RESULTS BY STREAM TYPE AT THE 10-S OUTPUT

| Family | Stream type | ARI | NMI | ACC | Exact K (%) | False sw./h |
|---|---|---|---|---|---|---|
| RC | Class-wise (5) | 0.949 | 0.935 | 0.980 | 100 | 0.00 |
| RC | Random duration (25) | 0.904 | 0.874 | 0.963 | 76.0 | 1.22 |
| NC | Class-wise (5) | 0.988 | 0.980 | 0.996 | 100 | 0.00 |
| NC | Random duration (25) | 0.957 | 0.934 | 0.984 | 100 | 0.06 |

The number of streams is given in parentheses.

with the correct number of environments in all 30 streams. The RC streams were harder than on TAU: at 10 s, the ARI was 0.876, with the correct count in 19 of the 30 streams (63.3%) and 2.4 false switches per hour. At 20 s, the RC ARI rose to 0.942, and the count was correct in 21 streams (70.0%).

As on TAU, the memory held exactly four environments at the end of every stream, and each of the 11 RC count errors at 10 s was a single provisional index in a random-duration stream; all five class-wise RC streams had the correct count (ARI 0.932). Unlike on TAU, 9 of these 11 provisional indices remained in the assignments issued at 20 s. Recognition remained nearly complete: at 10 s, 310 of 311 revisits were recognized in the RC streams and all 310 in the NC streams, and 87 (RC) and 89 (NC) of the 90 new environments were flagged. Times to the correct assignment were similar to those on TAU (live: 16.4 s in the RC streams and 7.4 s in the NC streams).

Two properties of CochlScene are likely to make its RC streams harder. Car and street form the least separable pair of either corpus, and because consecutive excerpts in an RC block usually come from different recordings, a block contains many transitions between recordings made by different participants with different phones, which can resemble changes of environment. On this corpus, the young-segment radius raised the number of RC streams with the correct count from 3 to 15, the learned vigilance raised it further to 19, compared with 15 when the vigilance selected on the development streams was used, and the revisit layer added 0.010 to the RC ARI. The three detector-and-memory

TABLE XIII
CONTRIBUTION OF THE COMPONENTS (10-S OUTPUT UNLESS STATED)

| Configuration | RC | | | | | NC | |
|---|---|---|---|---|---|---|---|
| | ARI | Exact K (%) | False sw./h | Live ARI | Exact K, 20 s (%) | ARI | False sw./h |
| Developed method | 0.912 | 80.0 | 1.02 | 0.806 | 96.7 | 0.962 | 0.05 |
| Without revisit layer | 0.861 | 80.0 | 1.30 | 0.744 | 96.7 | 0.923 | 0.47 |
| Without young-segment radius | 0.909 | 40.0 | 2.28 | 0.802 | 66.7 | 0.962 | 0.05 |
| Vigilance tuned on development data | 0.912 | 83.3 | 1.16 | 0.806 | 100 | 0.962 | 0.05 |
| Without the two learned thresholds | 0.910 | 33.3 | 2.49 | 0.802 | 70.0 | 0.962 | 0.05 |
| Detector and memory only | 0.859 | 33.3 | 2.77 | 0.740 | 70.0 | 0.922 | 0.47 |
| Detector and memory, RC vigilance | 0.774 | 93.3 (63.3) | 4.02 | 0.667 | 96.7 (70.0) | 0.925 | 0.61 |

Each row removes or replaces one or more components of the developed method. Development-tuned vigilance: $\rho = 0.0176$. Detector and memory only: development-tuned vigilance, no young-segment radius and no revisit layer. Without the two learned thresholds: development-tuned vigilance and no young-segment radius, with the revisit layer. RC vigilance: $\rho = 0.0216$, selected on the RC development streams only, with no young-segment radius and no revisit layer. The last three rows are the configurations of Section IV-L (BOCPD+ART2+layer; BOCPD+ART2; BOCPD+ART2, RC ρ). Values in parentheses (post hoc): identity-aware count (Section IV-J4), the percentage of streams with the correct count in which no two scenes were merged into one environment; shown only where it differs from exact K configurations of Table XIII (with the revisit layer, without it and with the RC vigilance) reached RC ARIs of 0.874, 0.864 and 0.830 on this corpus, against 0.876 for the developed method, and the correct count in 10.0%, 10.0% and 56.7% of the RC streams; with the RC vigilance, the identity-aware count was 40.0%, and the memory was correct in only 25 of the 30 streams.

### 6) TEST STREAMS OF TUT, SINS AND HEAR-DS

Table XV lists the post hoc results on physically separate test recordings from the development corpora; these streams were examined during method development. Over the 36 RC streams, the 10-s output reached an ARI of 0.929, with the correct count in 30 streams (83.3%) and 1.9 false switches per

TABLE XIV
PERFORMANCE OF THE DEVELOPED METHOD ON THE COCHLSCENE TEST STREAMS

| Family | Output | ARI | NMI | ACC | Exact K (%) | False sw./h |
|---|---|---|---|---|---|---|
| RC | Live | 0.754 | 0.718 | 0.899 | 63.3 | 3.59 |
| | 10 s | 0.876 | 0.846 | 0.950 | 63.3 | 2.40 |
| | 20 s | 0.942 | 0.922 | 0.977 | 70.0 | 2.17 |
| NC | Live | 0.883 | 0.845 | 0.955 | 100 | 0.19 |
| | 10 s | 0.969 | 0.950 | 0.988 | 100 | 0.05 |
| | 20 s | 0.981 | 0.967 | 0.993 | 100 | 0.00 |

Mean over the 30 streams of each family (5 class-wise and 25 random-duration). Exact K: percentage of streams in which the number of environments in the issued assignments equals the true number (four). Memory K was correct in all 60 streams.

hour; the 36 NC streams reached an ARI of 0.970, with the correct count in all of them. The results were highest on HEAR-DS (RC ARI 0.970 and 0.973), whose classes were defined for hearing aid environment recognition, and lowest on the RC streams of TUT-5 (0.794) and SINS-3 (0.921).

Unlike on TAU and CochlScene, the memory was not always correct. In three SINS-3 RC streams, one class was split into two environments (in two of them, a provisional index was also issued), and in two TUT-5 RC streams, two scenes were merged; in the latter, the issued count was nevertheless correct because a provisional index was also issued. The remaining three count errors were single provisional indices, as on TAU and CochlScene. Compared with the detector and memory alone, the developed method raised the number of RC streams with the correct count from 19 to 30 and the RC ARI from 0.901 to 0.929; when only the revisit layer was added to the detector and memory, the RC ARI was also 0.929, but the count was correct in 19 streams.

TABLE XV
RESULTS ON THE TUT, SINS AND HEAR-DS TEST STREAMS (10-S OUTPUT)

| Class set | Family | ARI | NMI | ACC | Exact K | False sw./h |
|---|---|---|---|---|---|---|
| TUT-3 | RC | 0.958 | 0.928 | 0.986 | 5/6 | 0.23 |
| | NC | 0.980 | 0.960 | 0.993 | 6/6 | 0.00 |
| TUT-4 | RC | 0.958 | 0.936 | 0.984 | 6/6 | 0.94 |
| | NC | 0.975 | 0.957 | 0.990 | 6/6 | 0.00 |
| TUT-5 | RC | 0.794 | 0.825 | 0.886 | 5/6 | 5.96 |
| | NC | 0.933 | 0.924 | 0.973 | 6/6 | 1.17 |
| SINS-3 | RC | 0.921 | 0.893 | 0.956 | 3/6 | 4.28 |
| | NC | 0.962 | 0.934 | 0.987 | 6/6 | 0.00 |
| HEAR-DS sep5 | RC | 0.973 | 0.959 | 0.989 | 5/6 | 0.24 |
| | NC | 0.986 | 0.976 | 0.994 | 6/6 | 0.00 |
| HEAR-DS noise4 | RC | 0.970 | 0.953 | 0.989 | 6/6 | 0.00 |
| | NC | 0.985 | 0.971 | 0.994 | 6/6 | 0.00 |
| All sets | RC | 0.929 | 0.916 | 0.965 | 30/36 | 1.94 |
| | NC | 0.970 | 0.954 | 0.989 | 36/36 | 0.19 |

Six streams per class set and family (one class-wise and five random-duration). Exact K: streams with the correct number of environments.

## L. COMPARISON WITH OTHER METHODS

The developed method was compared with configurations of its own detector and memory and with online methods from speaker diarization and environment analysis, all applied to the same CLAP embeddings and evaluated on the same streams and outputs. The detector-and-memory configuration (BOCPD+ART2) combines the detector of Section IV-C with the classifier of [52] applied to detected segments, using a fixed vigilance selected on the development streams; it was also evaluated with the revisit layer of Section IV-G (BOCPD+ART2+layer) and with a vigilance selected on the RC development streams only (BOCPD+ART2, RC ρ). The results of these three configurations are given in Table XIII; they are the comparators of the registered analysis below. A single-module CLAP adaptation of the app classifier [52] was also applied every second (ART2 per second), with the recorded buffering and comparison conventions; this does not reproduce the complete native two-module feature/VAD/fusion pipeline. The other methods are ΔBIC change detection with BIC clustering [66], the real-time segmentation and tracking method in [70], GLR change detection with incremental cosine clustering [69], cosine leader-follower clustering as in diart [91], online Bayesian clustering with a PLDA back-end [56] and a SoundSense-style model [54]. These are adaptations to a common representation: in particular, the diart comparator uses incremental cosine clustering rather than its complete overlap-aware diarization pipeline, and the Bayesian comparator uses an isotropic PLDA backend. The settings of every method were selected on the development streams of TUT, SINS and HEAR-DS.

Table XVI compares the 10-s outputs. On the RC streams, the developed method achieved the highest ARI among the compared methods on both corpora, 0.912 on TAU and 0.876 on CochlScene, against at most 0.794 and 0.734 for the other methods, both by SoundSense. The methods that decide every second (ART2 per second, leader-follower, PLDA clustering and SoundSense) produced 69 to 464 false switches per hour on the RC streams. On the NC streams, where the windows of each block are shuffled, several methods performed well; SoundSense reached a higher ARI than the developed method on both corpora (0.980 and 0.983 versus 0.962 and 0.969), as did PLDA clustering on CochlScene (0.972), while the developed method kept the correct count in every stream.

A primary analysis compared the developed method with the three detector-and-memory configurations (Table XIII) on the 30 RC streams at the 10-s output, in a fixed order (with the revisit layer, without it, and with the RC vigilance), stopping at the first comparison that failed [92]. A comparison passed if the developed method obtained the correct count in significantly more streams (one-sided exact sign test, $p < 0.05$) and its ARI was non-inferior with a margin of 0.02, that is, the lower limit of the 90% percentile bootstrap interval of the mean difference, computed by resampling the 30 paired stream-level ARI differences with replacement 20,000 times and taking the 5th and 95th percentiles [93], was above -0.02. The same procedure was applied to CochlScene as a separate replication. Table XVII shows that the comparisons with BOCPD+ART2+layer and BOCPD+ART2 passed on both corpora and that the sequence stopped at the RC vigilance, against which the count was not better (0 better and 4 worse streams on TAU; 6 better and 4 worse on CochlScene), although the ARI was higher by 0.137 and 0.046. Because the count can also be correct when a merge of two scenes is offset by an extra index, the counts were also scored post hoc with

TABLE XVI

COMPARISON WITH OTHER METHODS AT THE 10-S OUTPUT

| Method | TAU | | | | | CochlScene | | | | |
|---|---|---|---|---|---|---|---|---|---|---|
| | RC streams | | | NC streams | | RC streams | | | NC streams | |
| | ARI | Exact K (%) | False sw./h | ARI | Exact K (%) | ARI | Exact K (%) | False sw./h | ARI | Exact K (%) |
| Developed | 0.912 | 80.0 | 1.02 | 0.962 | 100 | 0.876 | 63.3 | 2.40 | 0.969 | 100 |
| ART2 per second [52] | 0.761 | 20.0 | 147.4 | 0.840 | 70.0 (66.7) | 0.507 | 0.0 | 175.5 | 0.957 | 96.7 |
| ΔBIC + BIC [66] | 0.786 | 76.7 | 4.00 | 0.925 | 100 | 0.447 | 60.0 (20.0) | 11.0 | 0.778 | 96.7 |
| Lu and Zhang [70] | 0.711 | 43.3 (33.3) | 3.34 | 0.738 | 70.0 | 0.588 | 56.7 (50.0) | 2.41 | 0.638 | 73.3 (63.3) |
| GLR + cosine [69] | 0.655 | 80.0 (73.3) | 6.04 | 0.744 | 96.7 (90.0) | 0.509 | 46.7 (43.3) | 10.9 | 0.694 | 100 (93.3) |
| Leader-follower [91] | 0.646 | 0.0 | 292.8 | 0.955 | 100 | 0.394 | 0.0 | 464.4 | 0.882 | 0.0 |
| PLDA clustering [56] | 0.730 | 3.3 | 198.2 | 0.911 | 90.0 (86.7) | 0.533 | 0.0 | 362.5 | 0.972 | 0.0 |
| SoundSense [54] | 0.794 | 3.3 | 69.2 | 0.980 | 100 | 0.734 | 0.0 | 109.7 | 0.983 | 90.0 |

Mean over 30 streams per corpus and family. Exact K: percentage of streams with the correct number of environments. Values in parentheses (post hoc): identity-aware count (Section IV-J4), the percentage of streams with the correct count in which no two scenes were merged into one environment; shown only where it differs from exact K.

the identity-aware count (Tables XIII and XVI, in parentheses). The developed method absorbed no scene in any TAU or CochlScene stream, so its counts were unchanged, whereas 9 of the 28 correct counts of BOCPD+ART2 with the RC vigilance on TAU and 5 of its 17 on CochlScene contained a merge. With the identity-aware count, the developed method was more effective than this configuration in 8 and less effective in 3 of the TAU streams (one-sided exact sign test, $p = 0.11$) and more effective in 8 and less effective in 1 of the CochlScene streams ($p = 0.020$).

On TAU test streams, two-sided sign-flip tests with Holm correction across the ten other methods [94] showed that the developed method achieved higher ARI than every other method (corrected $p \leq 0.012$), and obtained the correct environment count significantly more often than seven of them; the differences from ΔBIC + BIC and GLR + cosine were not significant, and BOCPD+ART2 with the RC vigilance obtained the correct count more often, also not significantly. On the test streams of TUT, SINS and HEAR-DS (Section IV-K6), BOCPD+ART2+layer matched the RC ARI of the developed method (0.929) but obtained the correct count in 19 instead of 30 of the 36 streams, as did BOCPD+ART2 (RC ARI 0.901), and the remaining methods reached an RC ARI of at most 0.880 (ART2 per second, with 55 false switches per hour).

TABLE XVII

COMPARISONS OF ENVIRONMENT-COUNT ACCURACY AND ARI ON RC TEST STREAMS (10-S OUTPUT)

| Compared with | Exact K: better / equal / worse | p | ARI difference (90% CI) | Result |
|---|---|---|---|---|
| *TAU (confirmatory)* | | | | |
| BOCPD+ART2+layer | 14 / 16 / 0 | $6.1\times10^{-5}$ | +0.002 (+0.001, +0.003) | Pass |
| BOCPD+ART2 | 14 / 16 / 0 | $6.1\times10^{-5}$ | +0.052 (+0.043, +0.061) | Pass |
| BOCPD+ART2, RC ρ | 0 / 26 / 4 | 1.00 | +0.137 (+0.105, +0.172) | Stop |
| *CochlScene (replication)* | | | | |
| BOCPD+ART2+layer | 16 / 14 / 0 | $1.5\times10^{-5}$ | +0.001 (-0.005, +0.007) | Pass |
| BOCPD+ART2 | 16 / 14 / 0 | $1.5\times10^{-5}$ | +0.012 (+0.006, +0.017) | Pass |
| BOCPD+ART2, RC ρ | 6 / 20 / 4 | 0.38 | +0.046 (+0.020, +0.075) | Stop |

Comparisons tested in the order listed; testing stops at the first that does not pass. p: one-sided exact sign test on exact K over the 30 streams. Pass: $p < 0.05$ and lower limit of the ARI interval above -0.02. The streams of a corpus reuse the same test recordings, so the p-values are conditional on these recordings.

## M. PARAMETER SENSITIVITY

Figure 5(a) illustrates why the vigilance is learned rather than fixed. On the TAU test streams, a fixed vigilance in the detector-and-memory configuration was usable only between about 0.014 and 0.022. Smaller values issued extra environments in every RC stream, and larger values merged scenes until the memory collapsed. Within this range, the value selected on the TUT, SINS and HEAR-DS development streams (0.0176) kept every scene separate but gave the correct count in only 10 of the 30 RC streams, whereas 0.0216 raised the correct count to 28 streams but merged scenes in nine of them, where an extra provisional index offset the merge; its memory was correct in only 21 streams, and the ARI fell to 0.774. The learned vigilance, in contrast, kept the memory correct in every TAU test stream without any such choice, and on CochlScene the developed method obtained the correct count in 19 RC streams, compared with 15 with the development-tuned vigilance. On the RC test streams of TUT, SINS and HEAR-DS, however, the memory became incorrect in five streams, and that of the detector-and-memory configuration, with the development-tuned vigilance, in four streams.

Figure 5(b) reports the sensitivity of the final method on the TAU test streams, with one fixed setting varied at a time around its frozen value. Among the varied settings, only the quantile used for the young-segment deviation changed the number of environments: with the 25th, 50th (frozen) and 75th percentiles, the count was correct in 20, 24 and 27 of the 30 RC streams at 10 s. No tested revisit-layer setting changed the issued count in these runs. Over the layer settings, the RC ARI ranged from 0.864 to 0.925. It changed little for thresholds up to 0.3 and slacks up to 0.16, increased steadily with the gate, from 0.881 at 0.045 to 0.925 at 0.125, and approached the value obtained without the layer (0.861) for thresholds of 2 or more, for which its contribution is small in these results. On the NC streams, the count was correct in every stream for every setting, and the ARI ranged from 0.923 to 0.969.

The detector settings were examined in the same way with the detector-and-memory configuration, which uses the same detector, on the test streams of Section IV-J5. The hazard rate (1/100 to 1/10,000), the alarm probability (0.8 to 0.99) and the merge interval (0 to 10 s) produced small aggregate differences (RC ARI at 20 s remained approximately 0.942, averaged over TUT, SINS and HEAR-DS with equal weight per corpus). The prior parameter $\kappa_0$ mainly affected the live output: from $\kappa_0$ = 0.001 to 1, the RC live ARI ranged from 0.774 to 0.847, while the ARI at 20 s stayed between 0.942 and 0.945. The prior noise scale $\beta_0/\alpha_0$ and the shape $\alpha_0$ were the sensitive settings: the RC ARI at 20 s was 0.731, 0.907, 0.942, 0.360 and 0.026 for $\beta_0/\alpha_0$ = 0.001, 0.003, 0.01 (frozen), 0.033 and 0.1, and $\alpha_0$ = 10 instead of 1 severely degraded the RC ARI at 20 s to 0.090. Because the embeddings are normalized, this scale is related to the representation rather than to a particular corpus, and the frozen value was used unchanged for all five corpora.

Sensitivity checks on the TUT, SINS and HEAR-DS development streams covered every value in Table IX except the encoder settings, the window length, the hop and the output delay. These checks assessed robustness and did not reselect any value. On TUT, SINS and HEAR-DS, the frozen method was close to the best development objective $J$ among the values tested. Among the values selected on development data, only the prior scale $\beta_0$ and the shape $\alpha_0$ changed the results substantially. Among the design values, the smoothing length $M = 3$ was the best of 1, 2, 3, 5 and 7 windows, the weight ½ in (13) was better than 0.3, 0.4, 0.6 and 0.7, the radius (15) was better than the sum $\rho + e(n)$ (RC ARI 0.946 against 0.854), and the distance (9) was better than the cosine distance and equal to the Euclidean distance in ARI. The median in (14) is a neutral choice: the 75th percentile gave the correct number of environments in more RC streams (84% against 79%) but more false switches.

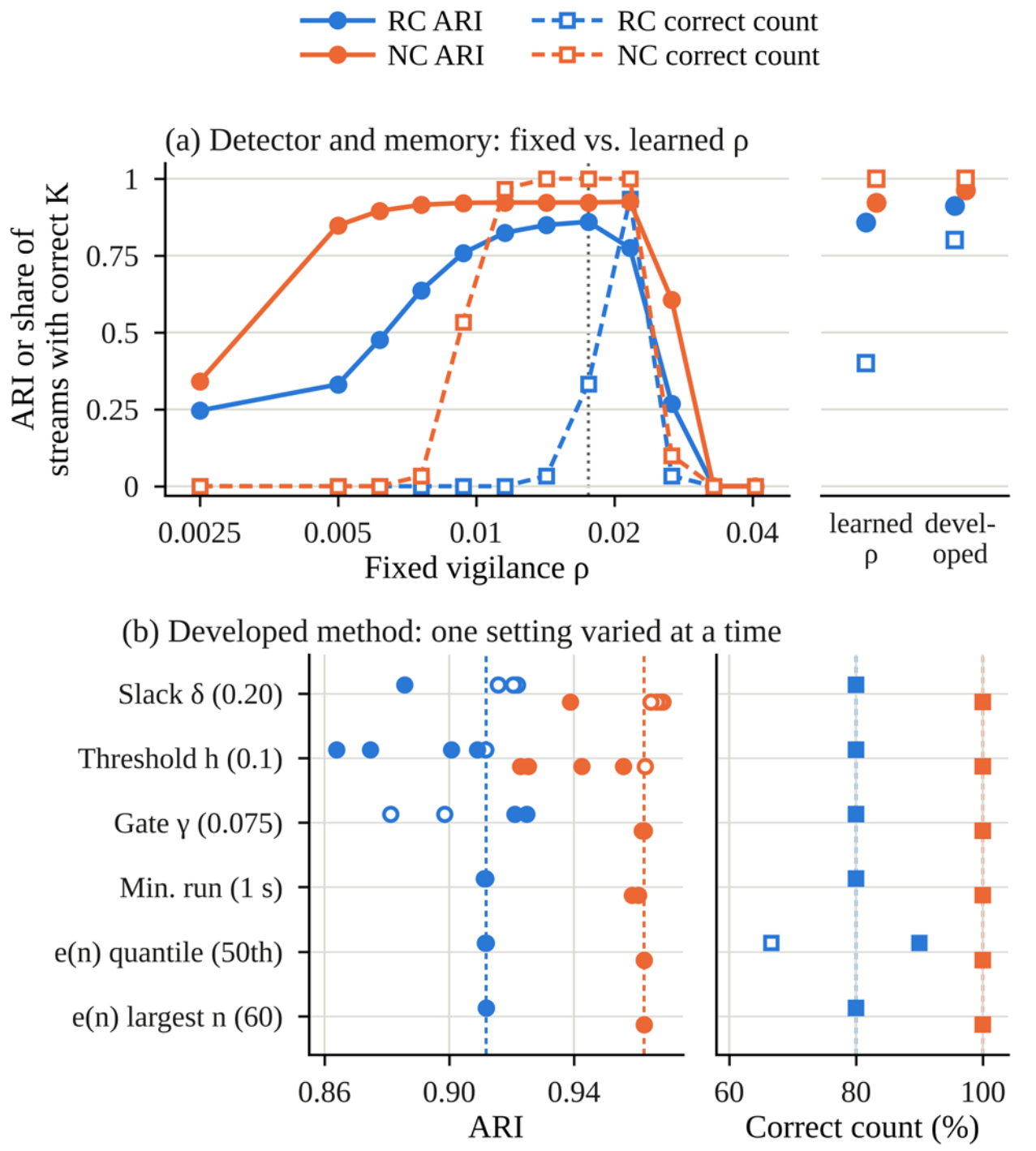

**FIGURE 5. Parameter sensitivity on the TAU test streams (10-s output; 30 RC and 30 NC streams). (a) Detector-and-memory configuration with a fixed vigilance ρ (dotted line: 0.0176, selected on TUT, SINS and HEAR-DS development streams) and, at the right, with the learned vigilance and for the developed method. (b) Developed method with one setting varied at a time around its frozen value (in brackets). Correct count: share of streams with the correct number of environments.**

## V. ENVIRONMENT-ADAPTIVE SWITCHING OF PERSONALIZED GAINS

This section combines the foreground VAD of Section III and the environment discovery of Section IV into the switching mechanism of Fig. 1 and evaluates it on continuous streams that contain speech and changing background environments. Importantly, the developed mechanism is not intended to adjust the gain in response to every minor change in the environment; rather, gain adaptation is considered only when the listening environments are sufficiently different to justify a meaningful change in the personalized gain setting or profile.

### A. ENVIRONMENT-SPECIFIC PERSONALIZATION GAINS

In the intended application, each discovered environment is associated with a gain profile and an indication of whether personalization has been completed. If a personalized gain profile is available, its gains are selected. Otherwise, the paired-comparison procedure of [14] would be initiated to obtain the user's preferred gains for that environment, with Bayesian preference learning carried out across the frequency bands. The default prescription would remain in use until personalization is completed. The resulting gain profile would then be stored with the environment identity and recalled when that environment returns.

For an environment $e$, the resulting personalized gain profile can be represented as $\mathbf{G}_e^* = [G_{e,1}^*, G_{e,2}^*, \dots, G_{e,N_{\text{band}}}^*]$, where $N_{band}$ denotes the number of frequency bands. Repeating the personalization procedure across four environments therefore results in a set of environment-specific personalized gain profiles, $\mathcal{P} = \{\mathbf{G}_1^*, \mathbf{G}_2^*, \mathbf{G}_3^*, \mathbf{G}_4^*\}$. Accordingly, the developed switching mechanism treats personalization separate from environmental analysis. The personalization stage determines the preferred gain profile for a particular audio environment, whereas the environment-discovery stage determines when the listening condition has changed sufficiently to warrant selection of a different personalized gain profile.

### B. INTEGRATED OPERATION

The foreground VAD of Section III identifies the audio frames that do not contain the designated talker; only these frames are passed to the environment discovery of Section IV, and each environment that it discovers is associated with its own gain profile. The aim is to demonstrate that a profile is activated for each newly encountered environment, that the same profile is recalled when an environment returns, and that switching follows the changes of the background rather than the speech of the talker. Both components are used exactly as specified in Sections III and IV; no parameter was selected or adjusted on the streams used here. Two conditions are examined on held-out recordings: a returning environment is either a later part of the same recording, as when the user returns to the same place, or a different recording of the same class, as when the user enters another place of the same type.

For scoring, the different-recording condition assumes a shared reference profile for each class; this broader transfer does not imply that every real-world recording within a semantic scene class should necessarily share one personalized gain profile. The same class-based scoring rule is retained for all reported results.

*Speech-free frame selection:* The VAD processes the 16-kHz mixture and issues a decision $\hat{y}_t \in \{0,1\}$ for every 20-ms frame *t*, with the bounded lookahead of 0.84 s described in Section III-G. Frames with $\hat{y}_t = 1$ are attributed to the designated talker and are withheld from environment analysis. All other frames, including pauses, non-speech sounds and speech not attributed to the designated talker, are treated as background: their 48-kHz samples are appended, in time order and without crossfading, to a background buffer. The environment discovery therefore operates on background time, which advances only while $\hat{y}_t = 0$, and the seconds of Section IV become background seconds rather than clock seconds.

*Embedding of each accumulated background second:* Each time one further full second of background audio has accumulated in the buffer, the most recent 2 s of the buffer are encoded with LAION-CLAP and smoothed with (6), giving the embedding $x_r$ of background second *r*. This is the representation of Section IV-B, with the 2-s window and 1-s hop counted in background time. The decision for the last frame of this window is available only after the lookahead, so $x_r$ becomes available at clock time

$$a(r) = \tau_r + 0.84 \text{ s} \tag{19}$$

where $\tau_r$ is the clock time at the end of the last frame of the window. While the VAD classifies incoming frames as foreground, the buffer does not grow, no embedding is produced, and the state of the environment discovery remains unchanged.

To evaluate switching independently of personalization, each environment-associated gain profile is represented by an identifier and is assumed to be available when required. The experiment measures whether the corresponding identifier is selected when an environment is encountered and recalled when that environment returns. The time required for user comparisons and gain estimation in [14] is outside this evaluation.

*Environment discovery and gain selection:* The embeddings are processed by the method of Section IV with the output delay $L = 10$ background seconds. The assignment $y_r^{(L)}$ of background second $r$ is issued once $L$ further embeddings have been received, at clock time $a(r + L)$. When an environment index is issued for the first time, the system treats it as a newly discovered environment and associates a personalized gain profile with it, so the *k*-th discovered environment receives the profile $G_k$. When an index that has been issued before is issued again, its profile is recalled. The active gain at clock time t is therefore the profile of the most recently issued assignment,

$$g(t) = G_{y_{r*}^{(L)}}, \quad r* = \max\{r : a(r + L) \le t\} \tag{20}$$

and the default prescription $G_0$ is used before the first assignment is issued.

### C. DEMONSTRATION STREAMS

Background audio was taken from the held-out test recordings of four HEAR-DS [87] classes representing clearly different listening conditions: cocktail party (C), in traffic (T), in vehicle (V) and wind turbulence (W), i.e., the noise4 set of Section IV. The test partition contains two or three recording situations per class, of 850-6042 s each. The front-left channel was used at 48 kHz.

The designated talker of each stream is one LibriSpeech test-clean speaker [76] with at least 8 min of speech; the five talkers comprise three female and two male speakers. Whole utterances are placed in random order without repetition and are separated by pauses drawn uniformly between 10 and 25 s. When required, all pauses of a stream are scaled by a common factor so that utterances occupy 28%-32% of the stream; the achieved proportion was 28.5%-31.9%. This proportion leaves extended speech-free intervals in every environment while requiring the environment analysis to bridge frequent interruptions by speech.

TABLE XVIII
DESIGN OF THE INTEGRATED OPERATION STREAMS

| Item | Setting |
|---|---|
| Background | HEAR-DS test recordings: C, T, V, W (2-3 per class) |
| Designated talker | One LibriSpeech test-clean speaker per stream (3 F, 2 M) |
| Speech share | 28.5%-31.9% of each stream |
| SNR | 0, 5 and 10 dB (same background and timing) |
| Visits | 8 per stream; each class twice, never consecutively |
| Visit duration | First 150-160 s; others 80-110 s (planned non-utterance time) |
| Changes during speech | 3 of 7 per stream |
| Returning visit | Same recording (≥ 60 s later) or different recording |
| Runs | 5 streams × 3 SNRs per condition; 17.9-20.9 min each |
| True changes | 105 per condition (45 to a new class, 60 returns) |

Planned non-utterance time excludes complete inserted utterance spans; accumulated background time instead follows the VAD-selected audio (Section V-B). C, cocktail party; T, in traffic; V, in vehicle; W, wind turbulence.

Each stream contains eight visits, two to each class, with no class visited twice in succession. Audio scene visit durations are specified in time outside complete utterances, termed planned non-utterance time. This construction quantity differs from the accumulated background time of Section V-B, which depends on the VAD decisions. The first visit lasts 150-160 s and the remaining visits last 80-110 s on this planning clock. These durations were chosen to provide opportunities for learning from the accumulated background audio: the young-segment deviation *e(n)* is recorded up to *n* = 60 only from segments of at least 120 windows, the within-segment statistic *W* requires 40 windows, the between-segment statistic *B* and the revisit layer require segments of at least 20 s, and the visits of the development streams lasted at least 80 s. In three of the seven changes of each stream, the background changes while the talker is speaking. Such a change is placed at a random point inside an utterance, and the preceding pause is shortened so that the previous visit keeps its planned non-utterance duration.

These durations define the operating conditions of the demonstration, not a minimum dwell time: no assignment waits for a segment to reach 120 background seconds, and a shorter completed segment contributes *e(n)* for smaller *n*. Shorter first visits and visits shorter than 80 s of planned non-utterance time were not tested as they do not constitute sustained audio backgrounds of concern to hearing aid users.

Every stream was built in two versions that differ only in the audio of the second visit to each class. In the same-recording condition, the second visit plays another part of the recording situation used for the first visit, starting at least 60 s after the end of the first visit's audio. In the different-recording condition, the second visit starts on another recording situation of the class. Each visit takes consecutive audio from a randomly drawn position, no background audio is repeated within a stream, and consecutive visits are joined with a 50-ms raised-cosine crossfade. The talkers, speech, visit boundaries and first visits are identical in both versions, and in every stream the mixture up to the first returning visit is identical sample by sample, so the two conditions can be compared change by change.

Speech is mixed at 0, 5 and 10 dB SNR using utterance-wise speech scaling: each utterance is scaled so that its strict-speech power relative to the background power over the utterance span equals the target value. The speech was scaled once, in the different-recording version, and reused unchanged in the same-recording version, so the achieved SNR of utterances in returning visits can deviate slightly from the target. A single clipping-prevention factor is applied per stream, so the three SNR versions of a stream share identical background audio, timing and reference labels. Five streams at three SNRs give 15 runs per condition, of 17.9-20.9 min (4.89 h per condition), each condition containing 105 true environment changes: 45 to a class not yet visited in the stream and 60 returns to a class visited before. Recording situations recur across the five streams. Table XVIII summarizes the design.

### D. EVALUATION PROCEDURE

The reference gain, called the expected gain, is derived from the ground truth by numbering the true environments in the order of their first appearance in the stream: the first environment corresponds to $G_1$, the next new environment to $G_2$, and a returning environment keeps its number. The selected gain is the profile identifier activated by the system, which assigns a number to its profiles by the order of discovery without access to the ground truth. Expected-gain correctness is scored by comparing this selected identifier with the ground-truth identifier. When environments are merged or discovered late, matching identifiers need not indicate that a selected profile belongs to the true environment.

A second reference assesses recall of the profile associated with each true environment. During the second half of that environment's first visit, the profile active for the greatest total duration is selected; an equal-duration tie is resolved by its earliest occurrence within that half. It is accepted as the owned-profile reference only if it is a nondefault profile created during that first visit. If the selected profile is the default G0 or was created before that visit, no owned-profile reference is established. This reference remains fixed for later visits; an environment without one is scored as unsuccessful under this reference. Using the second half is a scoring convention intended to allow for the initial switching delay. This reference is assigned retrospectively for evaluation and is not supplied to the online system.

For every true change, the first gain activated after the change, whether it is the expected gain, and the number of other gains activated before the expected one are recorded, together with the time to the expected gain, i.e., the clock time from the true change until the expected gain becomes active. This time includes every stage of the system: the remainder of an utterance during which the change occurs, the accumulation of background audio, the VAD lookahead, the confirmation of the change and the output delay. A change is also counted as kept when its first gain is the expected gain and no other gain is activated before the visit ends, and the gain active at the end of each visit is recorded. Reaching the expected gain before the visit ends and having it active at the end is scored separately, because a correct activation can be followed by another switch. For returns, recall is scored against the owned-profile reference, and the creation of any additional profile is recorded separately. In addition, the number of profiles created in each run and all additional gain switches within a true visit after its first switch are counted. The initial activation at the start of the stream is excluded from this additional-switch count. The VAD is evaluated with the frame measures of Section III-H, together with the proportion of reference speech in the background seconds passed to the environment discovery (leaked speech). For this percentage, leaked reference-speech time excludes transition collars, while the denominator is all retained background time. For the different-recording condition, every return is also characterized by its distance to the stored environment (Section V-E5).

## E. OVERALL RESULTS

### 1) SPEECH-FREE FRAME SELECTION

Table XIX summarizes the VAD in both conditions, which share the same speech. In the same-recording condition, the macro-F1 was 89.9%, 93.5% and 94.6% at 0, 5 and 10 dB, and at most 2.7% of the background frames were withheld as foreground speech. Recall was the lowest for cocktail party, which contains competing talkers, and for in traffic (56.7% and 59.4% at 0 dB; 85.9% and 88.3% at 10 dB). Speech of the designated talker missed by the VAD therefore reached the environment discovery mainly in these classes: at 0 dB, reference speech outside transition collars formed 13.0% of the cocktail-party and 10.4% of the in-traffic background seconds, and 7.1% of all background seconds. The different-recording condition gave similar values, and every visit in either condition retained at least 82 background seconds.

### 2) GAIN SELECTION WITH RETURNS FROM THE SAME RECORDING

Table XX reports the gain selection in both conditions. In the same-recording condition, the first gain activated after each of the 105 changes was the expected gain, and no other gain was activated in between, and the expected gain then remained active until the visit ended. Each of the 60 returns recalled the gain profile stored for its environment, and each run created exactly four gain profiles, one for each class, so no environment was split into two profiles or merged with another. Every run contained exactly eight gain switches: the initial activation of $G_1$ about 13 s after the start of the stream and one switch for each of the seven changes. No gain switch occurred without a true change, including at the onsets and ends of utterances, and the results were the same at the three SNRs despite the lower VAD recall at 0 dB. The detector confirmed all 105 changes without additional alarms and placed each change within 3 background seconds of the true boundary.

TABLE XIX
FOREGROUND VAD ON THE INTEGRATED OPERATION STREAMS (%)

| Return, SNR | Precision | Recall | Macro-F1 | FFR | C recall | T recall | Leaked speech |
|---|---|---|---|---|---|---|---|
| Same, 0 dB | 93.1 | 77.2 | 89.9 | 1.91 | 56.7 | 59.4 | 7.1 |
| Same, 5 dB | 92.7 | 87.6 | 93.5 | 2.31 | 76.1 | 81.1 | 4.0 |
| Same, 10 dB | 91.9 | 92.1 | 94.6 | 2.73 | 85.9 | 88.3 | 2.6 |
| Different, 0 dB | 92.9 | 75.5 | 89.2 | 1.93 | 50.2 | 58.1 | 7.6 |
| Different, 5 dB | 92.7 | 88.6 | 93.8 | 2.34 | 79.3 | 81.1 | 3.7 |
| Different, 10 dB | 91.8 | 93.9 | 95.2 | 2.80 | 91.7 | 89.1 | 2.0 |

Same/Different: returning visits from the same or a different recording. Precision and recall of the speech class, pooled over the five streams. FFR: false-foreground rate (background frames classified as speech). C, T recall: recall in cocktail party and in traffic. Leaked speech: reference-speech time outside transition collars in the retained audio, divided by all background time passed to environment discovery.

The expected gain became active within 30 s of the change in 87.6% of the changes and within 60 s in all of them, with a median of 17.5 s (17.3, 17.6 and 18.0 s at 0, 5 and 10 dB); the slowest switch took 40.0 s. Over the total stream time, the expected gain was active 87.2% of the time; the remaining time consists of the start-up interval and of the intervals between each change and the corresponding switch, during which the previous profile remained active. Although some foreground speech remained in the retained audio (7.1% of the background seconds at 0 dB), the first profile activated after every true change was correct, so correct profile selection did not require perfect rejection of foreground speech.

### 3) EXAMPLE OF GAIN SWITCHING

Figure 6 shows the operation of the complete system on stream 1 of the same-recording condition at 5 dB, and Table XXI lists its seven changes. In Fig. 6(a), the expected gain is derived from the ground truth, whereas the selected gain is the profile activated by the system. The system activates $G_1$ for wind turbulence at 12.8 s, creates $G_2$, $G_3$ and $G_4$ when cocktail party, in vehicle and in traffic first appear, and recalls $G_2$, $G_3$, $G_1$ and $G_4$ when these environments return. Each switch follows its change after 13.5-34.0 s, and the selected gain does not change in response to any of the utterances of the talker.

Figure 6(b) shows the first appearance of in traffic at 614.2 s, which occurs while the talker is speaking. The VAD withholds most of the utterance, so few background seconds become available until it ends. The detector places the change at the background second that becomes available at 619.7 s and raises an alarm 3 background seconds later, at 622.7 s. When the assignment of that background second is issued, $L =$ 10 background seconds after it became available, the new profile $G_4$ is activated at 629.7 s, 15.5 s after the change. The

TABLE XX
GAIN SELECTION AFTER THE 105 TRUE ENVIRONMENT CHANGES OF EACH CONDITION (10-S OUTPUT)

| **Measure** | **Same recording** | **Different recording** |
|---|---|---|
| First gain after change, matches expected (%) | 100 | 86.7 |
| Expected gain reached before the visit ended (%) | 100 | 97.1 |
| Expected gain active at the end of the visit (%) | 100 | 91.4 |
| Expected gain selected first and kept to the end of the visit (%) | 100 | 81.0 |
| Returns recalling their owned-profile reference (%) | 100 | 95.0 |
| Profiles created per run (true: 4) | 4 in every run | 5.1 (4-6) |
| Wrong switches per hour | 0 | 4.9 |
| Additional switches within a visit after its first switch | 0 | 25 |
| Expected gain within 15 s (%) | 27.6 | 20.0 |
| Expected gain within 30 s (%) | 87.6 | 76.2 |
| Expected gain within 60 s (%) | 100 | 97.1 |
| Median time to expected gain (s) | 17.5 | 20.9 |
| Longest time to expected gain (s) | 40.0 | 49.5 |
| Time with expected gain active (%) | 87.2 | 81.4 |

15 runs per condition (5 streams at 0, 5 and 10 dB). Times are clock times measured from the true change; the longest time is over the changes whose expected gain was reached. Recall and wrong-switch counts use the owned-profile reference; the remaining correctness and timing measures use the expected-gain reference. These references coincide for both HEAR-DS conditions in this table. Wrong switches activate a profile other than the owned-profile reference; in the different-recording condition, 20 of the 24 activated an additional profile created for a return; the other 4 activated a profile of another class (Section V-E5). Time with expected gain active includes the start-up interval and the interval after each change.

profile $G_2$ of cocktail party remains active until then, so the speech at the change does not cause any intermediate switch.

### 4) ORIGIN OF THE SWITCHING DELAY

The switching delay is determined mainly by the output delay. Because $L$ is counted in background seconds, the 10-s wait becomes longer in clock time when the talker speaks during it, and the accumulation of background audio and the VAD lookahead add to it. In six of the seven changes in Table XXI, the gain switched $L$ or $L + 1$ background seconds after the background second at which the detector placed the change became available. In change 5, the detector raised its alarm 27 background seconds after the background second at which it placed the change, and the gain switched at the alarm. In change 3, a return, the stored profile was recalled before the detector raised its alarm, through the revisit layer of Section IV-G. Over all the runs, the median alarm delay measured from the true background boundary was 10 background seconds (range 2-30), and the median time to the correct gain was 17.3 s for changes during speech and 18.3 s for changes in pauses. Delays above 30 s occurred in 13 of the 105 changes, ten of which were changes to a new class; a new environment cannot be recognized by the revisit layer, so the switch to a new gain profile must wait for the detector. The live output of the same method switched after a median of 8.3 s, also without any wrong switch, and the 20-s output after 29.6 s.

### 5) RETURNS FROM A DIFFERENT RECORDING

In the different-recording condition, the first gain activated after a change was the expected gain in 86.7% of the changes, and the expected gain was reached before the end of the visit in 97.1% (Table XX). The 45 changes to a new class were handled as in the same-recording condition. The errors concerned returns: in 16 of the 60 returns, an additional profile was created during the visit, so that 12 of the 15 runs contained one or two additional profiles. The stored profile was the first gain activated in 46 of the 60 returns, and 3 never recalled it. No gain profile was owned by two classes, although in two runs (stream 5 at 0 and 10 dB), profiles of another class were active for 13-18 s after two changes in each run; in one of these two returns per run, the first decision on the returning segment nevertheless recalled the stored profile (Table XXII).

To examine why, each return was characterized by the distance of Section IV-D between its embeddings and the stored environment (Table XXII). On the stored embeddings of the same recordings without speech, in 1000 random draws of a 155-window first visit and a 95-window return per class, a return from a different recording lay 1.9 times farther from the stored environment than a return from the same recording, although the stored environment of its own class remained the nearest in 99.4% of the draws. In the streams, the mean embedding of a same-recording return lay at a median of 0.36 of the matching radius in use when the return was first decided,

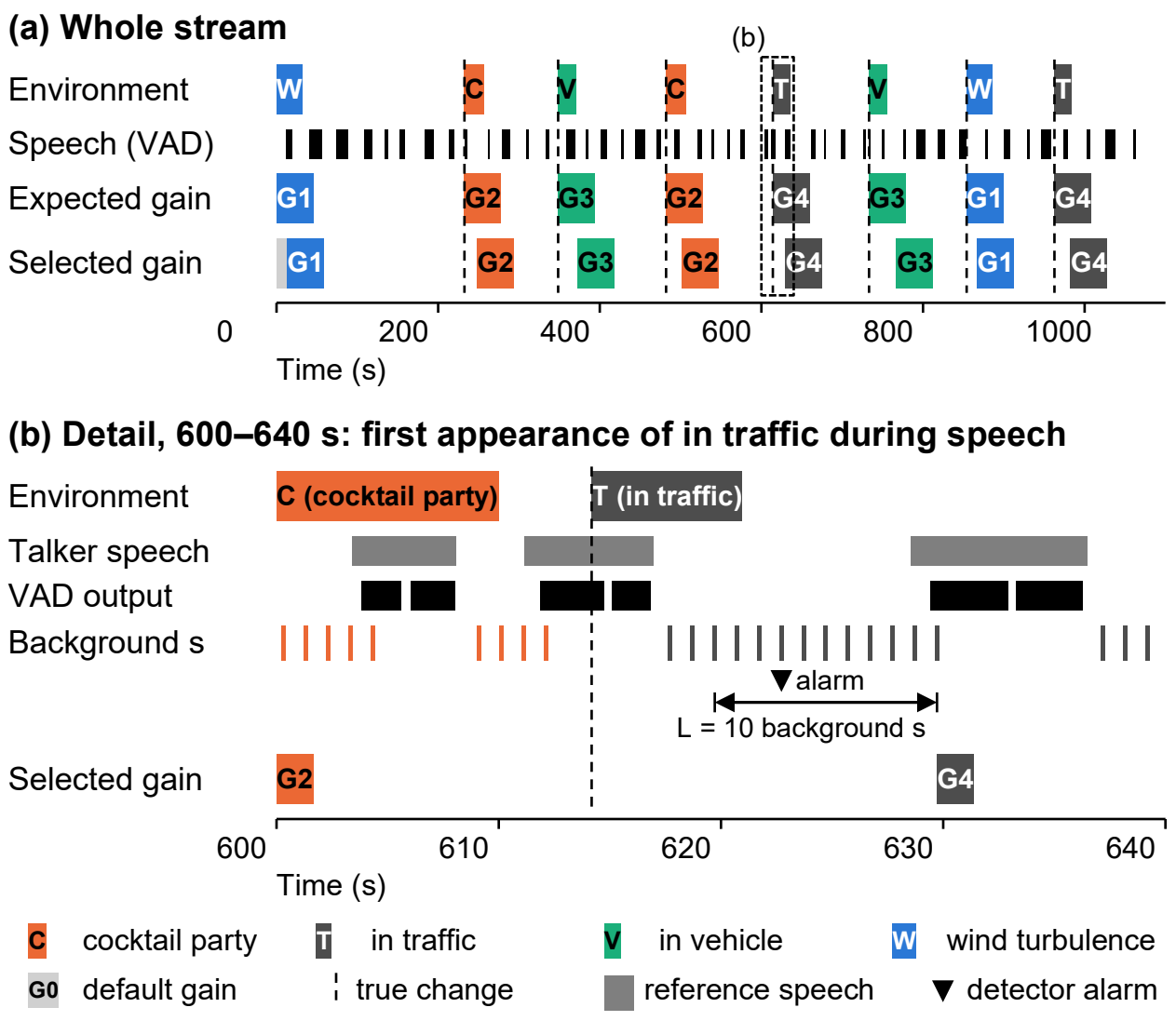

**FIGURE 6. Operation of the integrated switching system on stream 1 at 5 dB SNR, with returns from the same recording. (a) Whole stream: true environment, foreground speech detected by the VAD, expected gain derived from the ground truth, and gain selected by the system from the assignments issued at a 10-s delay. (b) Detail of 600-640 s: first appearance of in traffic during speech. Ticks mark the clock times at which background seconds become available to the environment discovery, colored by their true environment; the bracket marks the output delay of the background second at which the change is placed. G0: default prescription.**

TABLE XXI
GAIN SWITCHING IN STREAM 1 AT 5 DB (RETURNS FROM THE SAME RECORDING)

| No. | Change (s) | From → to | Type | Expected | Selected | Switch (s) | Delay (s) |
|---|---|---|---|---|---|---|---|
| 1 | 232.5* | W → C | New | G2 | G2 | 248.5 | 16.0 |
| 2 | 348.5 | C → V | New | G3 | G3 | 372.2 | 23.7 |
| 3 | 482.1 | V → C | Return | G2 | G2 | 501.8 | 19.7 |
| 4 | 614.2* | C → T | New | G4 | G4 | 629.7 | 15.5 |
| 5 | 733.2* | T → V | Return | G3 | G3 | 767.2 | 34.0 |
| 6 | 854.0 | V → W | Return | G1 | G1 | 867.5 | 13.5 |
| 7 | 962.6 | W → T | Return | G4 | G4 | 982.6 | 19.9 |

*: change during speech. Expected: profile identifier assigned by the ground-truth order of first appearance. Selected: profile identifier activated by the system. Delay: time from the true change to the switch to the expected gain. G1 was activated for W at 12.8 s.

TABLE XXII
DISTANCE OF RETURNING ENVIRONMENTS TO THE STORED ENVIRONMENT HEAR-DS TEST RECORDINGS

| Measure | Same recording | Different recording |
|---|---|---|
| Without speech: median distance to own class | 0.0079 | 0.0148 |
| Without speech: own class nearest (%) | 99.9 | 99.4 |
| Without speech: within learned vigilance (%) | 98.8 | 88.2 |
| In streams: distance / radius, median (maximum) | 0.36 (0.58) | 0.67 (1.04) |
| In streams: recalled at the first decision | 60 of 60 | 48 of 60 |
| In streams: not recalled, whole return within radius | 0 | 9 |
| In streams: not recalled, beyond radius | 0 | 3 |
| In streams: nearer to another class | 0 | 0 |

Distance: mean absolute difference (Section IV-D). Without speech: stored embeddings of the test recordings, 1000 draws per class and condition (4000 per condition across four classes); learned vigilance: median final ρ of the different-recording runs (0.020). In streams: mean embedding of each return, compared with the stored environment and the matching radius in use at the first decision on the returning segment (all SNRs). Descriptive analysis, specified after the different-recording results were known; it was not used to choose any setting. The first decision on the returning segment in this table differs from the first gain switch after the true change reported in Section V-E5: 48 of 60 versus 46 of 60 different-recording returns, respectively.

and at most at 0.58, whereas a different-recording return lay at a median of 0.67, and at most at 1.04. Thus 54 of the 60 different-recording returns lay within the radius as a whole, but with a smaller margin. Because the first assignment of a returning segment is issued after only 10 background seconds (Section IV-H), when its mean is less stable than over the whole visit, this first decision exceeded the radius in 9 returns, and in three further returns the whole visit lay beyond it; the remaining three returns beyond the radius were recalled at the first decision. None of the returns was nearer to the environment of another class.

### F. LIMITATIONS OF THE EXPERIMENTATIONS

In these streams, the integrated operation of the switching system converted the VAD decisions into a background stream from which every environment was discovered when it first appeared and, when the same recording returned, recognized every time without a single wrong switch. The experimentations nevertheless have several limitations. Recognition of a returning environment depends on the vigilance learned from the stream, which reflects the variation observed within and between the segments encountered so far. When an environment is first observed in a single recording, a different recording of the same type of place lies closer to the edge of the learned tolerance, and a short initial segment can exceed it, so that an additional profile is created. On held-out TAU recordings, whose scenes are less distinct than the HEAR-DS classes, the same different-recording construction reached the owned-profile reference defined in Section V-D before the end of the visit after 77.1% of the changes (81 of 105), and returns were also confused with other scenes. Under the expected-gain numbering defined in Section V-D, the corresponding rate was 87.6% (92 of 105). Learning the tolerance from observations that represent the variation of each environment, such as several recordings of it, is left for future work. Each class has only two or three test recording situations, so recordings recur across the five streams. The profiles are assumed to be available as soon as an environment is discovered, whereas the procedure of Section V-A requires paired comparisons in that environment; until these are completed, a newly discovered environment would have to use the default or another existing profile. Each stream contains a single designated talker without conversational turn-taking, and the talkers come from LibriSpeech test-clean, which was not used to fit or calibrate the VAD but was used in its evaluation in Section III. Finally, the streams were processed offline with causal processing and bounded lookahead. The demonstrated operating scope in the experimentations reported is therefore gain profile selection among a limited set of distinguishable, sustained environments in streams with one designated foreground speaker.

## VI. CONCLUSION

This paper has provided a switching mechanism or system for environment-adaptive personalized gains in hearing aids. A one-class foreground voice activity detector selects the audio that does not contain a talker; an online environment discovery method analyzes this audio to detect changes of the background environment, to recognize environments that return and to discover new ones without a predefined number of environments; and the gain profile personalized for the identified environment is then selected automatically.

The voice activity detector was trained only on speech mixed with noise and combines a fixed speech-in-noise reference bank with a speaker-consistency follow-up, so that background speech such as babble is retained as part of the environment. Across 36 background types, most of which were excluded from fitting and calibration, it achieved three-seed mean macro-F1 values of 85.53%-91.98%. Its three-seed mean macro-F1 exceeded every reported MFCC ORVAE and HuBERT VAE decision rule for all eight speech/background corpus pairs. The environment discovery combines Bayesian online change-point detection with a segment-level memory whose two thresholds, namely the vigilance and the young-segment deviation, are learned from the incoming stream, and with a gated CUSUM layer that recognizes returning environments. On the TAU test streams, its assignments issued with a 10-s delay reached adjusted Rand indices of 0.91 and 0.96 for the recording-continuous and non-contiguous streams, with the correct number of environments in 80.0% and 100% of the streams. With unchanged settings, it reached recording-continuous ARI of 0.876 on CochlScene.

In a held-out experimentation on continuous streams containing speech and four audio environments commonly encountered by hearing aid users, where a returning environment was a later part of the same recording, the correct gain profile was the first one activated after each of the 105 environment changes, every return recalled its stored profile, no wrong profile was activated, and the correct profile became active after a median of 17.5 s. When a returning environment was a different recording of its class, the correct gain profile was reached before the end of the visit after 97.1% of the changes and was active when the visit ended after 91.4%, but an additional gain profile was created for 16 of the 60 returns.

The evaluation also showed the main limitations. Recognizing a different recording of a recurring environment remains harder than detecting a change: on CochlScene, the number of environments was correct in only 63.3% of the recording-continuous streams at the 10-s output. The switching delay is dominated by this output delay, and the demonstration assumed that a gain profile is available as soon as an environment is discovered.

Our future work will implement the entire switching system in real-time on a smartphone for field deployment purposes.

## ACKNOWLEDGMENT

The authors acknowledge the computational resources provided by the University of Texas at Dallas. OpenAI's ChatGPT and Anthropic's Claude were used to assist with reviewing the grammar and readability.